\documentclass[journal,onecolumn]{IEEEtran}
 \usepackage{times}

\usepackage{algorithm}
\usepackage{algorithmic}

\usepackage{graphics} 
\usepackage{epsfig} 
\usepackage{mathrsfs}
\usepackage{times} 
\usepackage{amsmath} 

\usepackage{amsthm}
\usepackage{amssymb}  
\usepackage{gensymb}

\usepackage[footnotesize]{caption} 

\usepackage{float,bm}
\usepackage{wasysym}
\usepackage[normalem]{ulem}
\usepackage{mathtools}
\usepackage{bbm}
\usepackage{graphicx}
\usepackage{caption}
\usepackage{subcaption}
\usepackage{color}  
\usepackage[dvipsnames]{xcolor}
\usepackage{titlesec}
\usepackage{indentfirst}
\usepackage{multirow}

\usepackage{perl_acronyms}
\usepackage{perl_misc}

\usepackage{color, colortbl}

\usepackage{soul} 

\usepackage{geometry}
\usepackage[
    style=ieee,
    doi=false,
    isbn=false,
    url=false,
    eprint=false,
    backend=bibtex,
    natbib=true
    ]{biblatex}

 \usepackage{nopageno}

\bibliography{references}

\usepackage{hyperref}
\newtheorem{defn}{Definition}

\newtheorem{lem}[defn]{Lemma}

\newtheorem{assum}[defn]{Assumption}

\newtheorem{thm}[defn]{Theorem}

\definecolor{init_traj_blue}{HTML}{5577FE}
\definecolor{final_traj_green}{HTML}{1DB032}
\definecolor{init_pose_turq}{HTML}{3FC8D1}
\definecolor{final_pose_gold}{HTML}{D8A621}
\definecolor{obs_red}{HTML}{C80000}

\providecommand{\methodname}{\text{PHLAME}}

\newcommand{\stkout}[1]{{\color{Orange}\ifmmode\text{\sout{\ensuremath{#1}}}\else\sout{#1}\fi}}

\providecommand{\X}{x}

\providecommand{\A}{\mathcal{A}}

\providecommand{\J}{\mathcal{J}}

\providecommand{\OO}{\mathcal{O}}
\providecommand{\R}{\ensuremath \mathbb{R}}
\providecommand{\N}{\ensuremath \mathbb{N}}

\providecommand{\psX}{\xi}

\providecommand{\Jpi}{J_{\Xi_i}}

\providecommand{\uu}{u}
\providecommand{\x}{\mathtt{x}}
\providecommand{\xd}{\dot{\mathtt{x}} }

\providecommand{\Xd}{\dot \X}
\providecommand{\Xdd}{\ddot \X}
\providecommand{\Fd}{F_d}

\providecommand{\Fdx}{F_d(\x(t))}
\providecommand{\Fx}{F(\x(t))}

\providecommand{\q}{q}
\providecommand{\qd}{\dot{q}}
\providecommand{\qdd}{\ddot{q}}
\providecommand{\Hq}{H(q(t))}
\providecommand{\Hinvq}{H^{-1}(\x_{P1}(t))}
\providecommand{\Hinv}{H^{-1}}
\providecommand{\Cq}{C(\x_{P1}(t),\x_{P2}(t))}

\providecommand{\Ginv}{G^{-1}}
\providecommand{\dGdx}{\frac{\partial G}{\partial \X}}
\providecommand{\dFddx}{\frac{\partial \Fd}{\partial \X}}
\providecommand{\dFdTdx}{\frac{\partial \Fd^T}{\partial \X}}

\providecommand{\Act}{\mathcal{A}}
\providecommand{\dxds}{\frac{\partial \X}{\partial s}}

\providecommand{\ddt}{\frac{d}{dt}}
\providecommand{\dldx}{\frac{\partial L}{\partial \X}}
\providecommand{\dldxd}{\frac{\partial L}{\partial \dot{\X}}}
\providecommand{\dldxs}{\frac{\partial L}{\partial \X_s}}
\providecommand{\dldxds}{\frac{\partial L}{\partial \dot{\X}_s}}

\providecommand{\xpone}{\X_{P1}}
\providecommand{\xptwo}{\X_{P2}}
\providecommand{\xdpone}{{\dot \X}_{P1}}
\providecommand{\xdptwo}{{\dot \X}_{P2}}
\providecommand{\xddpone}{{\ddot \X}_{P1}}
\providecommand{\xddptwo}{{\ddot \X}_{P2}}

\providecommand{\Hdot}{{\dot H}}
\providecommand{\HdotT}{{\dot H^T}}
\providecommand{\HT}{{H^T}}
\providecommand{\Cdot}{{\dot C}}

\providecommand{\FDuzero}{FD_{0}}

\providecommand{\dmHinvCdxpone}{ \frac{\partial \FDuzero}{\partial \xpone} }
\providecommand{\dmHinvCdxptwo}{ \frac{\partial \FDuzero}{\partial\xptwo} }
\providecommand{\dFDdxpone}{ \frac{\partial \FDuzero}{\partial \xpone} }
\providecommand{\dFDdxptwo}{ \frac{\partial \FDuzero}{\partial \xptwo} }
\providecommand{\dFDTdxpone}{ \frac{\partial \FDuzero^T}{\partial \xpone} }
\providecommand{\dFDTdxptwo}{ \frac{\partial \FDuzero^T}{\partial\xptwo} }

\providecommand{\ddFDddxpone}{\frac{\partial^2 FD_0}{\partial \xpone^2}}
\providecommand{\ddFDddxptwo}{\frac{\partial^2 FD_0}{\partial \xptwo^2}}
\providecommand{\ddFDdxponetwo}{\frac{\partial^2 FD_0}{\partial \xpone \partial \xptwo}}

\providecommand{\dHTdxpone}{\frac{\partial \HT}{\partial \xpone}}

\providecommand{\dHTHdx}{\frac{\partial (\HT H)}{\partial \X}}
\providecommand{\dHTHdxi}{\frac{\partial (\HT H)}{\partial \X_i}}
\providecommand{\dHTHdxone}{\frac{\partial (\HT H)}{\partial \X_1}}
\providecommand{\dHTHdxN}{\frac{\partial (\HT H)}{\partial \X_{N}}}
\providecommand{\dHTHdxNone}{\frac{\partial (\HT H)}{\partial \X_{N+1}}}
\providecommand{\dHTHdxtwoN}{\frac{\partial (\HT H)}{\partial \X_{2N}}}
\providecommand{\dHTdxi}{\frac{\partial \HT}{\partial \X_{i}}}
\providecommand{\dHdxi}{\frac{\partial H}{\partial \X_{i}}}
\providecommand{\dHTHdxtwoN}{\frac{\partial (\HT H)}{\partial \X_{2N}}}

\providecommand{\dHdxpone}{\frac{\partial H}{\partial \xpone}}

\providecommand{\ddHddxpone}{\frac{\partial^2 H}{\partial \xpone^2}}
\providecommand{\dHdotdxpone}{\frac{\partial \Hdot}{\partial \xpone}}
\providecommand{\dHdotTdxpone}{\frac{\partial \Hdot^T}{\partial \xpone}}
\providecommand{\dCdotdxpone}{\frac{\partial \Cdot}{\partial \xpone}}
\providecommand{\dHdotdxdpone}{\frac{\partial \Hdot}{\partial \xdpone}}
\providecommand{\dHdotTdxdpone}{\frac{\partial \Hdot^T}{\partial \xdpone}}
\providecommand{\dCdotdxdpone}{\frac{\partial \Cdot}{\partial \xdpone}}
\providecommand{\dCdotdxdptwo}{\frac{\partial \Cdot}{\partial \xdptwo}}

\providecommand{\dbdxg}{\frac{\partial b(g_j(\X))}{\partial \X}}

\providecommand{\dCdxpone}{\frac{\partial C}{\partial \xpone}}
\providecommand{\dCdxptwo}{\frac{\partial C}{\partial \xptwo}}

\providecommand{\ddCddxpone}{\frac{\partial^2 C}{\partial \xpone^2}}
\providecommand{\ddCddxptwo}{\frac{\partial^2 C}{\partial \xptwo^2}}
\providecommand{\ddCdxponetwo}{\frac{\partial^2 C}{\partial \xpone \partial \xptwo}}
\providecommand{\ddCdxptwoone}{\frac{\partial^2 C}{\partial \xptwo \partial \xpone}}

\providecommand{\dCdotdxpone}{\frac{\partial \Cdot}{\partial \xpone}}
\providecommand{\dCdotdxptwo}{\frac{\partial \Cdot}{\partial \xptwo}}

\providecommand{\ddIDddxpone}{\frac{\partial^2 ID}{\partial \xpone^2}}
\providecommand{\ddIDddxptwo}{\frac{\partial^2 ID}{\partial \xptwo^2}}
\providecommand{\ddIDdxponetwo}{\frac{\partial^2 ID}{\partial \xpone \partial \xptwo}}

\providecommand{\FDuzero}{FD_{u=0}}

\providecommand{\dHTHdxpone}{\frac{\partial (\HT H)}{\partial \xpone}}

\providecommand{\dHTHinvdxpone}{\frac{\partial (\HT H)^{-1}}{\partial \xpone}}

\providecommand{\HTHinv}{(\HT H)^{-1}}
\providecommand{\dgammadxpone}{\frac{\partial \gamma}{\partial \xpone}}
\providecommand{\dgammadxptwo}{\frac{\partial \gamma}{\partial \xptwo}}
\providecommand{\dgammadxdpone}{\frac{\partial \gamma}{\partial \xdpone}}
\providecommand{\dgammadxdptwo}{\frac{\partial \gamma}{\partial \xdptwo}}

\providecommand{\dgammadxddptwo}{\frac{\partial \gamma}{\partial \xddptwo}}

\providecommand{\dalphaonedxpone}{\frac{\partial \alpha_1}{\partial \xpone}}
\providecommand{\dalphaonedxptwo}{\frac{\partial \alpha_1}{\partial \xptwo}}
\providecommand{\dalphatwodxpone}{\frac{\partial \alpha_2}{\partial \xpone}}
\providecommand{\dalphatwodxptwo}{\frac{\partial \alpha_2}{\partial \xptwo}}

\providecommand{\dalphaonedxdptwo}{\frac{\partial \alpha_1}{\:\partial \xdptwo}}

\providecommand{\dalphatwodxdptwo}{\frac{\partial \alpha_2}{\partial \xdptwo}}
\providecommand{\dGammadxpone}{\frac{\partial \Gamma}{\partial \xpone}}
\providecommand{\dGammadxptwo}{\frac{\partial \Gamma}{\partial \xptwo}}

\providecommand{\dGammadxdptwo}{\frac{\partial \Gamma}{\partial \xdptwo}}

\providecommand{\dOmegatwodx}{\frac{d \Omega_2}{d \X}}
\providecommand{\dOmegathreedx}{\frac{d \Omega_3}{d \X}}
\providecommand{\dOmegafourdx}{\frac{d \Omega_4}{d \X}}

\providecommand{\dOmegatwodxd}{\frac{d \Omega_2}{d \Xd}}
\providecommand{\dOmegathreedxd}{\frac{d \Omega_3}{d \Xd}}
\providecommand{\dOmegafourdxd}{\frac{d \Omega_4}{d \Xd}}

\providecommand{\dxpone}{\frac{d}{d \xpone}}
\providecommand{\dxptwo}{\frac{d}{d \xptwo}}
\providecommand{\dxpone}{\frac{d}{d \xpone}}
\providecommand{\dxptwo}{\frac{d}{d \xptwo}}

\providecommand{\ccons}{c_{\text{cons}}}

\providecommand{\Ufb}{u_{fb}}

\title{\LARGE \bf Bring the Heat: Rapid Trajectory Optimization with Pseudospectral Techniques and the Affine Geometric Heat Flow Equation}
\author{Challen Enninful Adu$^1$, César E. Ramos Chuquiure$^1$, Bohao Zhang$^1$, and Ram Vasudevan$^1$
\thanks{$^{1}$Robotics, University of Michigan, Ann Arbor, MI. {\tt\small <enninful, cesarch, jimzhang, ramv>@umich.edu}}%
\thanks{This work was funded by MURI and the Automotive Research Center (ARC)}
}
\begin{document}

\setlength{\textfloatsep}{18pt}

\maketitle
 \vspace*{-0.5cm} 
\thispagestyle{empty}
\pagestyle{plain}

\begin{abstract}
Generating optimal trajectories for high-dimensional robotic systems in a time-efficient manner while adhering to constraints is a challenging task. 
To address this challenge, this paper introduces \methodname{}, which applies pseudospectral collocation and spatial vector algebra to efficiently solve the Affine Geometric Heat Flow (AGHF) Partial Differential Equation (PDE) for trajectory optimization. 
Unlike traditional PDE approaches like the Hamilton-Jacobi-Bellman (HJB) PDE, which solve for a function over the entire state space, computing a solution to the AGHF PDE scales more efficiently because its solution is defined over a two-dimensional domain, thereby avoiding the intractability of state-space scaling.
To solve the AGHF one usually applies the Method of Lines (MOL), which works by discretizing one variable of the AGHF PDE, effectively converting the PDE into a system of ordinary differential equations (ODEs) that can be solved using standard time-integration methods.
Though powerful, this method requires a fine discretization to generate accurate solutions and still requires evaluating the AGHF PDE which can be computationally expensive for high dimensional systems.
\methodname{} overcomes this deficiency by using a pseudospectral method, which reduces the number of function evaluations required to yield a high accuracy solution thereby allowing it to scale efficiently to high-dimensional robotic systems. 
To further increase computational speed, this paper presents analytical expressions for the AGHF and its Jacobian, both of which can be computed efficiently using rigid body dynamics algorithms.
The proposed method \methodname{} is tested across various dynamical systems, with and without obstacles and compared to a number of state-of-the-art techniques.
\methodname{} is able to generate trajectories for a 44-dimensional state-space system in $\sim5$ seconds, much faster than current state-of-the-art techniques.
A project page is available at \href{https://roahmlab.github.io/PHLAME/}{https://roahmlab.github.io/PHLAME}.

\end{abstract}

\begin{figure}[H]
    \centering
    \includegraphics[trim={0cm, 0cm, 0cm, 0cm},clip,width=1\columnwidth,angle=0]{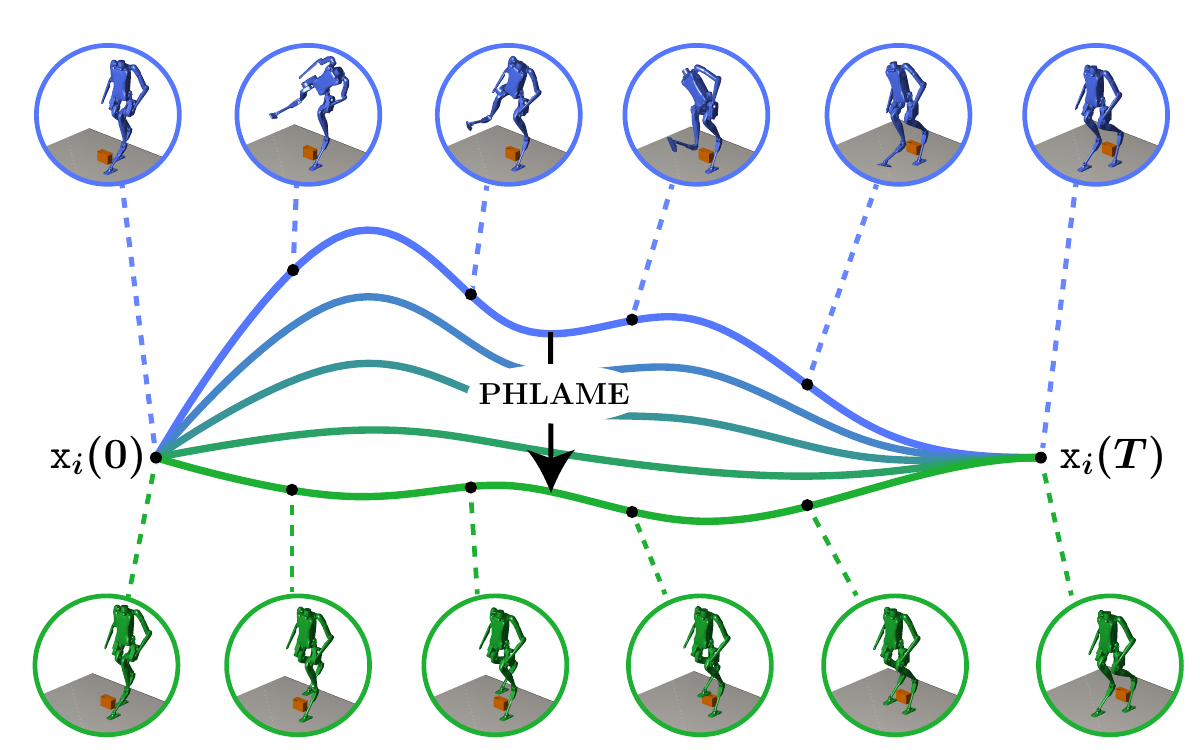}
    \caption{\methodname{} works by first taking in some initial guess of a trajectory (trajectory of $i$th state shown in \textcolor{init_traj_blue}{\textbf{dark blue}}) which does not have to be dynamically feasible and evolves it into some dynamically feasible final trajectory \textcolor{final_traj_green}{\textbf{dark green}}).
    Both trajectories start and end at $\x(0) = \x_0$ and $\x(T) = \x_f$ respectively.
    Notice that at the initial trajectory Digit (a high dimensional humanoid robot) has a dynamically infeasible set of configurations during it's stepping trajectory and that at the end of the \methodname{} solve that trajectory is made into a dynamically feasible one where Digit is able to step over the box.}
    \label{fig:AGHF-intro}
\end{figure}

\section{Introduction}\label{sec:intro}
To perform effectively in real-world applications, robots must generate dynamically feasible trajectories across a diverse range of tasks—including ground vehicle navigation, manipulation, and legged locomotion \cite{hong2022agile, ewen2021generating, michaux2024let, kim_cheetah3, liuradius, minicheetah}. Optimal control plays a fundamental role in enabling these capabilities, providing the mathematical framework necessary for planning and executing complex movements under various physical constraints. 
For robotic applications, an optimal control algorithm must satisfy several critical requirements: (1) computational efficiency to enable online planning and replanning, (2) scalability to handle high-dimensional systems like humanoids and manipulators, (3) ability to incorporate nonlinear dynamics and constraints for real-world tasks, and (4) reliable convergence with minimal sensitivity to initial conditions.
Despite significant advances in optimal control theory and algorithms, existing methods face fundamental challenges in simultaneously meeting these requirements. 
This paper addresses these challenges by proposing a novel algorithm that enhances computational efficiency while maintaining dynamic feasibility, leveraging recent advances in pseudospectral methods and spatial vector algebra applied to the Affine Geometric Heat Flow (AGHF) Partial Differential Equation (PDE).

Existing optimal control approaches have made significant theoretical and algorithmic advances, yet still struggle to simultaneously meet all these requirements. To understand these challenges, we examine the two primary approaches to optimal control: Dynamic Programming (DP) methods and Variational methods.

Dynamic programming methods \cite{bertsekas2012dynamic} leverage Bellman's Principle of Optimality to compute optimal value functions and policies. Global approaches that solve the Hamilton-Jacobi-Bellman (HJB) equation provide optimality guarantees regardless of initialization but face fundamental computational barriers. 
In continuous time, these methods require discretizing the entire state space, meaning that as the state dimension increases, the number of grid points this nonlinear PDE must be evaluated at scales exponentially.
This makes these methods computationally intractable beyond 5-dimensional systems and increasingly unstable numerically as dimensions grow.
To address these limitations, Differential Dynamic Programming (DDP) variants such as Crocoddyl \cite{crocoddyl2020} apply dynamic programming principles locally along a trajectory, achieving faster convergence through iterative forward-backward passes. 
While more computationally tractable than global methods, DDP approaches can struggle with convergence when initialized far from local optima and face challenges incorporating inequality constraints, though recent work like Aligator \cite{aligator} has begun addressing these limitations.

Variational methods, which derive necessary conditions for optimality using calculus of variations, offer an alternative approach. Direct methods in this category discretize the continuous optimal control problem into a nonlinear program (NLP).
For example, direct collocation methods like C-FROST \cite{C-FROST} and TROPIC \cite{Tropic2020} approximate trajectories using either linear interpolation, which is used in schemes such as trapezoidal collocation, or polynomial basis functions, which are employed in Hermite-Simpson collocation.
While these methods effectively handle complex constraints \cite{smit2017energetic,smit2019walking}, they require significant computation time for high-dimensional systems, often taking tens of minutes to converge for high dimensional systems. 
Moreover, the quality of their solutions can be sensitive to both discretization choices and initial trajectory guesses. 
Another direct method, RAPTOR \cite{zhang2024rapidrobusttrajectoryoptimization}, parameterizes trajectories using Bézier curves, which allows it to solve optimization problems in a matter of seconds and makes it more robust to initial guesses. 
However, RAPTOR does not explicitly enforce dynamic constraints, and as a result, requires the robotic system to be fully actuated.

Indirect variational methods like Pontryagin's Maximum Principle (PMP) \cite{pmp} provide elegant theoretical solutions but exhibit high sensitivity to initial guesses of co-states, making them challenging to apply to complex robotic systems where good initialization is difficult to obtain.

This paper presents a novel algorithm that addresses these limitations through an alternative PDE-based formulation called the Affine Geometric Heat Flow (AGHF).
 First introduced in \cite{AGHF_OG}, the AGHF poses trajectory generation as the solution to a PDE that evolves an initial trajectory that may not be dynamically feasible into a final trajectory that is dynamically feasible while minimizing control input magnitudes. 
 Unlike traditional PDE-based optimal control methods like HJB whose solution domain scales with the state space dimension, the AGHF solution has a two-dimensional domain regardless of system dimension.
As a result, the AGHF PDE offers a significant advancement in computational speed, without compromising the dynamic feasibility of motion planning and is also able to incorporate path constraints.
Because the AGHF solution has a two-dimensional domain, it is usually solved by using the Method of Lines (MOL) \cite{schiesser2012numerical}.
The MOL begins by discretizing the domain of the solution of AGHF along one dimension to generate a set of nodes.
Then at each node, it represents the PDE as if it is an Ordinary Differential Equation (ODE).
This system of ODEs can then be solved by using well understood numerical ODE solvers.

The nodes in the MOL are usually chosen in an evenly spaced fashion, and the solution quality gets better as more nodes are used.
The number of evaluations of the PDE function scales linearly with the number of nodes so having a fine grid requires many evaluations of the AGHF to compute a solution.
This issue is further exacerbated for high dimensional systems because evaluating the AGHF requires evaluating the dynamics and derivatives of the dynamics of the system whose trajectory is being optimized.
This has made applying the AGHF PDE to perform trajectory optimization untenable for high dimensional systems \cite{AGHF_OG}.

To address these challenges, this paper proposes \methodname{} which applies pseudospectral collocation in conjunction with spatial vector algebra to rapidly solve the AGHF PDE. 
The main contributions of this work are four-fold:
First, we propose a pseudospectral method that reduces the number of AGHF evaluations and nodes when compared to the classical MOL, which allows \methodname{} to scale up to high dimensional robotic systems (Section \ref{subsec: pseudospectral}).
Second, we provide an analytical expression for the AGHF in terms of the rigid body dynamics equation and an algorithm to rapidly evaluate this analytical AGHF expression using spatial vector algebra based rigid body dynamics algorithms. (Sections \ref{subsec: pinocchio pde computation} \ref{subsubsec: computation efficiency AGHF}).
Third, we provide an analytical expression for the jacobian of the AGHF and an algorithm to rapidly compute it using spatial vector algebra based rigid body dynamics algorithms (Sections \ref{subsec: jacobian} \ref{subsubsec: computation efficiency jacobian}).
Finally, this paper demonstrates the performance of \methodname{} for trajectory optimization for a number of different dynamical systems in the presence of obstacles and without obstacles and illustrates its performance when compared to a variety of state of the art methods (Section \ref{sec:experiments}).

The remainder of the paper is arranged as follows:
Section \ref{sec: notation} presents the background and introduces the relevant notation for the paper. 
Section \ref{sec:AGHF} introduces the AGHF and discusses the underlying theory associated with the AGHF. 
Section \ref{sec: AGHF constraints} details how to incorporate constraints into the AGHF to enable actions like obstacle avoidance.
\section{Preliminaries}
\label{sec: notation}
This section introduces the notation used throughout this manuscript.
This paper is focused on performing trajectory optimization for robot systems whose dynamics can be written as follows:
\begin{equation}
\label{eq: manipulator dyn}
    \Hq\qdd(t) + C(\q(t), \qd(t)) = B \uu(t),
\end{equation}
where $q(t) \in \R^{N}$ is the configuration of the robot at time $t$, $\uu(t) \in \R^{m}$ is the input applied to the robot at time $t$,
$\Hq$ is the mass matrix, $C(\q(t), \qd(t))$ is the grouped Coriolis and gravity term and $B$ is the actuation matrix.
For convenience, let $\x(t)$ correspond to the vector of $\q(t)$ and $\qd(t)$. 
To be consistent with the notation in the rest of the paper we refer to the first $N$ and last $N$ components of $\x(t)$ as $\x_{P1}(t)$ and $\x_{P2}(t)$, respectively. 
Additionally, let \textbf{0} be a $N\times1$ vector of zeros.
Using these definitions, we can represent the dynamics of the robot \eqref{eq: manipulator dyn} as a control affine system:
\begin{equation}
\label{eq: control affine dyn}
    \xd(t) = \Fdx + \Fx \uu(t),
\end{equation}
where 
\begin{align}
\Fdx &= \begin{bmatrix}
      \label{eqn: Fd}
       \x_{P2}(t) \\
        -\Hinvq\Cq 
    \end{bmatrix} \\
\Fx &= \begin{bmatrix}
        0_{N \times m} \\
        \Hinvq B 
    \end{bmatrix}
\end{align}

For convenience, we assume without any loss of generality that we are interested in the evolution of the system for $t \in [0,T]$.

To ensure the convergence of the AGHF PDE, we make the following assumption on the differentiability and smoothness of the system dynamics and existence of a feasible solution:
\begin{assum}\label{assum:dynamics continuity}
Both $F_d$ and $F$ are $C^2$, Lipschitz continuous, and $F$ has constant rank almost everywhere in $\R^n$.
Additionally, we assume the existence of a feasible solution to the motion planning problem for the system.
\end{assum}

\noindent Note that the dynamics of rigid body robotic systems are smooth and Lipschitz continuous when their domain is restricted to a compact set. 

The objective of this paper is to develop an algorithm to construct a trajectory beginning from some user-specified initial condition, $\x_0$, and ending in some user-specified terminal condition, $\x_f$, while avoiding obstacles and satisfying the dynamics in \eqref{eq: control affine dyn} for all $t \in [0,T]$ while minimizing the square control. 
If we let the zero superlevel set of a function $g$ represent the inequality constraints, then one can formulate the trajectory design problem as the solution to the following optimization problem:
\begin{equation*}
\begin{aligned}
& \underset{\uu \in L^2}{\text{inf}}
& & \int_{0}^{T} \| \uu(t)\|_2^2 ~dt &\quad \text{(OCP)} \\
& \text{s.t.}
& & \xd(t) = \Fdx + \Fx \uu(t), \quad &\forall t \in [0, T], \\
& & & g(\x(t)) \leq 0 \quad &\forall t \in [0, T], \\
& & & \x(0) = \x_0,& \\
& & & \x(T) = \x_f, &
\end{aligned}
\end{equation*}
where $L^2$ denotes the space of square integrable functions.
Note if a particular $\x(\cdot)$ satisfies each of the constraints in the (OCP), then we call $\x(\cdot)$ a feasible trajectory to (OCP).
To numerically solve this optimization problem, the methods discussed in Section \ref{sec:intro} are typically used. 
As mentioned in Section \ref{sec:intro}, these methods either have high computational costs, are highly sensitive to initial guesses, or may have difficulty dealing with non-smooth elements like obstacles or non-convex constraints. 
The proposed approach aims to address these challenges by enabling the rapid generation of trajectories for high-dimensional systems while incorporating multiple constraints using the Affine Geometric Heat Flow Partial Differential Equation to solve (OCP).
\section{The Affine Geometric Heat Flow (AGHF) Partial Differential Equation}
\label{sec:AGHF}
The Affine Geometric Heat Flow (AGHF) Partial Differential Equation (PDE) is a parabolic PDE that attempts to solve (OCP). 
At a high level, the AGHF equation works by deforming an initial trajectory that begins from $\x_0$ and ends at some final state $\x_f$ into a final trajectory that begins from $\x_0$ and ends at $\x_f$.
The AGHF deforms that initial trajectory, which can be any trajectory including one that does not satisfy the dynamics, into a dynamically feasible final trajectory that minimizes some user-specified cost. 
This section summarizes the background knowledge and theory of the AGHF.
A more detailed treatment of the subjects discussed here can be found in \cite{AGHF_OG,liu2019homotopy,liu2020geometric}.
Note throughout this section, we assume that there are no inequality constraints in (OCP). 
The inequality constraint case is considered in Section \ref{sec: AGHF constraints}. 

\subsection{Homotopies and Extracting Control Inputs}

To describe the evolution of trajectories by the AGHF PDE, we begin by defining a homotopy: $\X:[0,T] \times [0,s_{max}] \to \R^{2N}$, that is twice differentiable with respect to its first argument and differentiable with respect to its second argument. 
For convenience, we denote $\X(t,s)$ by $\X_s(t)$ and we denote $\frac{\partial \X}{\partial t}(t,s)$ by $\dot{\X}(t,s)$ or $\dot{\X}_s(t)$. 
Next, define the \emph{Lagrangian} as:
\begin{align}
\label{eqn:lagrangian}
    L(\X_s(t),\dot{\X}_s(t)) = \left(\dot{\X}_s(t) - F_d(\X_s(t))\right)^T G(\X_s(t)) \left(\dot{\X}_s(t)- F_d(\X_s(t))\right)
\end{align}
where $G: \R^{2N} \to \R^{2N \times 2N}$ is a user-specified matrix. 
Similar to in Section \ref{sec: notation}, for succinctness we refer to the first $N$ and last $N$ components of $\X_s$ as $\X_{P1}$ and $\X_{P2}$, respectively. 
Additionally, let the first and second time derivatives of these states be defined as $\xdpone$, $\xdptwo$ and $\xddpone$, $\xddptwo$ respectively.
In Section \ref{subsec:G}, we describe how to select $G$ to ensure that the AGHF minimizes the squared control effort as in the cost function in (OCP). 
Finally, we define the \emph{Action Functional}:
\begin{equation}
\label{eqn:action functional}
    \Act(\X_s) = \int_0^T  L(\X_s(t),\dot{\X}_s(t)) dt.
\end{equation}

Using these definitions, we can write down the AGHF PDE: 
\begin{defn}\label{defn:AGHF}
The \emph{Affine Geometric Heat Flow} is a parabolic partial differential equation that is defined as:
\begin{align}
\label{eqn:AGHF}
   \dxds(t,s)  = \Ginv(\X(t,s)) \bigg(\ddt \dldxds(\X_s(t),\dot{\X}_s(t)) - \dldxs(\X_s(t),\dot{\X}_s(t))\bigg)
\end{align}
with the following boundary conditions:
\begin{align}
\X_s(0) &= \x_0,  \quad \forall s \in [0,s_{max}] \\ 
\X_s(T) &= \x_f, \quad \forall s \in [0,s_{max}].
\end{align}
\end{defn}

When solving the AGHF PDE, one begins by specifying an initial curve $\X_{init}:[0,T] \to \R^{2N}$ and setting it such that $\X_0 = \X_{init}$.
As the AGHF PDE evolves forward in $s$, one can prove that the action functional is minimized. 
In addition, if during that evolution the AGHF converges to a curve where the right hand side of the AGHF PDE is equal to $0$, then one has found a curve that extremizes the action functional.
Such a curve is called a \emph{steady state solution}.
We formalize these observations in the following lemma that was originally proved in \cite[Lemma 1]{AGHF_OG} and which we repeat here for convenience:
\begin{lem}\label{lem:AGHF}
Let $\X$ satisfy the AGHF PDE. 
Then, $\frac{d \Act(\X_s)}{ds} \leq 0$ for all $s$.
In addition, if the right hand side of the AGHF PDE when evaluated at $\X_{s^*}$ is equal to $0$ for some $s^* \in [0,s_{max})$, then $\frac{d\Act(\X_{s^*})}{ds} = 0$. 
\end{lem}

\subsection{Ensuring $\Act$ Coincides with the Control Input by Designing $G$}
\label{subsec:G}

To ensure that the action functional being minimized coincides with minimizing the square of the control input as in (OCP), we must design $G$ carefully. 
We do this by applying the following lemma that was originally proven in \cite[Theorem 1]{AGHF_OG}:
\begin{lem}\label{lem:G}
Suppose (OCP) is feasible and let $G$ be defined as follows:
    \begin{equation}
    \label{eqn:G(x)}
    G(\X_s(t)) = (\bar{F}(\X_s(t))^{-1})^T K \bar{F}(\X_s(t))^{-1} 
    \end{equation}
where  
\begin{equation}
    \label{eqn: definition of K}
    K = \begin{bmatrix}
        kI_{N \times N} & 0_{N \times N} \\
        0_{N \times N} & I_{N \times N}
        \end{bmatrix} \in \R^{2N \times 2N}
\end{equation}

for $k > 0$ and
\begin{equation}
\label{eqn:fbar}
    \bar{F}(\X_s(t)) =  \begin{bmatrix}
        F_c(\X_s(t)) & F(\X_s(t))
        \end{bmatrix} \in \R^{2N \times 2N},
\end{equation}
where $F_c \in \R^{2N \times (2N-m)}$ is some differentiable in $\X$ matrix such that $\bar{F}$ is invertible for all $\X$ in $\R^{2N}$.
Note that such an $F_c$ can be obtained using the Gram-Schmidt procedure.
Let $\uu_s:[0,T] \to \R^{m}$ be the extracted control inputs at some $s \in [0, s_{max}]$ given by:
\begin{equation}
\label{eqn:control extraction}
    \uu_s(t) = \begin{bmatrix} 0_{N \times  N} & I_{N \times N} \end{bmatrix}
    \bar{F}(\X_s(t))^{-1} (\dot{\X}_s(t) - \Fd(\X_s(t))).
\end{equation}
Then 
\begin{equation}
\label{eqn:action functional k}
    \Act(\X_s) = \int_0^T k \|\xdpone - \xptwo\|_2^2 + \| \uu_s(t) \|_2^2 dt.
\end{equation}
\end{lem}
\noindent In short, at each $s$, the Action Functional with $G$ as described by Lemma \ref{lem:G} corresponds to the squared control input of the trajectory $\X_s$ plus the error between the velocity states ($\xptwo$) and the derivative of the position states ($\xdpone$).
For sufficiently large $k$, this penalizes errors in the dynamics, reducing them as the trajectory evolves.
For feasible trajectories of (OCP) that satisfy the dynamics, this error is zero, and the Action Functional with $G$ corresponds solely to the squared control input generating that feasible trajectory.

As a result, if $\X_s$ was a feasible trajectory of (OCP) for each $s$, then Lemmas \ref{lem:AGHF} and \ref{lem:G} would ensure that the AGHF was minimizing the square of the control input during its evolution.
Though we do not describe it here, \cite[Theorem 1]{AGHF_OG} under Assumption \ref{assum:dynamics continuity} proves that for sufficiently large $k$ and $s_{max}$, the control extracted from the solution to the AGHF PDE can be used to generate a trajectory that is arbitrarily close to a feasible trajectory of (OCP).
In fact \cite[Theorem 1]{AGHF_OG} proves an explicit bound on how close the trajectory generated by using \eqref{eqn:control extraction} is to a feasible trajectory of (OCP).

\section{Solving the AGHF rapidly for high dimensional systems}
\label{sec:solving AGHF}

The computationally intensive part of the AGHF method is solving \eqref{eqn:AGHF}, which is a parabolic PDE.
In contrast to traditional PDEs used for optimal control (e.g., the Hamilton-Jacobi-Bellman PDE), the AGHF PDE has a complexity that scales polynomially with increasing state dimension rather than exponentially.
The favorable scaling properties of the AGHF are owed to the fact that the domain of \eqref{eqn:AGHF} is always \textbf{two-dimensional} and the dimension of the range of the function scales \textbf{linearly} with the state dimension.
However, evolving the AGHF quickly demands being able to evaluate the right hand side of \eqref{eqn:AGHF} rapidly. 
This can be difficult for high dimensional systems as the system dynamics and its derivatives must be evaluated each time the AGHF is called.
If the AGHF must be evaluated at many time nodes, as in the classical MOL algorithm, then it can be even more challenging to construct a technique to rapidly solve the AGHF PDE. 

This section describes how our method for solving the AGHF addresses these issues.
Section \ref{subsec: pinocchio pde computation} describes how to leverage spatial vector algebra to rapidly compute the right hand side of \eqref{eqn:AGHF}.
Section \ref{subsec: pseudospectral} describes how to apply a pseudospectral MOL to reduce the number of time nodes that need to be considered to generate an accurate solution.
Notably this pseudospectral MOL approach also allows us to accurately compute derivatives of time.
This enables our method to avoid having to compute derivatives using finite difference, which dramatically reduces the number of function evaluations.

\subsection{Computing AGHF PDE Partial Derivatives Analytically}
\label{subsec: pinocchio pde computation}

This subsection derives analytical expressions that can be used to evaluate \eqref{eqn:AGHF} in terms of the rigid body dynamics equation \eqref{eq: manipulator dyn}, and how to leverage spatial vector algebra to rapidly compute these expressions.
We summarize the relevant results in the following theorem whose proof can be found in Appendix \ref{sec: AGHF RHS Appendix}.

Note, for succinctness, we have dropped the dependence on $(t,s)$ for the following equations (i.e. $\X(t,s)$ is  denoted by $\X$).
Additionally, in similar fashion to the notation introduced in Section \ref{sec: notation}, we denote the first $N$ and last $N$ components of $\X$ as $\X_{P1}$ and $\X_{P2}$, respectively. 
Lastly, we also drop the dependence on the $\X$ terms for the dynamics functions~(i.e., $H(\xpone)$ is denoted by just $H$ and so on)

\begin{thm}
\label{thm: Analytical PDE expression}
    Consider a system with dynamics as in \eqref{eq: manipulator dyn}. 
    The AGHF PDE \eqref{eqn:AGHF} using the $G$ described in Lemma \ref{lem:G} can be written as follows:
    \begin{equation}
    \label{eqn:aghf omega}
    \begin{split}
        & \dxds = \Omega \bigg( \X, \dot{\X}, \ddot{\X}, k \bigg) = \Omega_1 - (\Omega_2 - \Omega_3 + \Omega_4),
    \end{split}
    \end{equation}
    where 
    \begin{equation}
    \label{eqn: ginv_dt_dl_dxd}
    \begin{split} 
        \Omega_1 =
        2 \begin{bmatrix}  \Xdd_{P1} - \Xd_{P2} \\ 
            (\HT H)^{-1} \big((\HdotT H + \HT \Hdot) \Xd_{P2} + \HT H \Xdd_{P2} + \HdotT C + \HT \Cdot \big)
        \end{bmatrix}
    \end{split}
    \end{equation}
    
    \begin{align}
    \label{eqn: omega2}
    \begin{split}
         \Omega_2 = \begin{bmatrix} -\frac{1}{k}I_{N \times N} \dFDTdxpone \HT \bigg(2C + 2H \dot{\X}_{P2} \bigg) \\ 
            -(\HT H)^{-1} \dFDTdxptwo \HT \bigg(2C + 2H \dot{\X}_{P2} \bigg) \bigg) 
        \end{bmatrix} \\
    \end{split}
    \end{align}

    \begin{equation}
        \label{eqn: omega3}
        \Omega_3 =
        \begin{bmatrix}
            \vspace{3.1pt}
            \textbf{0} \\
            \vspace{3.1pt}
             2k (\HT H)^{-1} (\dot{\X}_{P1} -{\X}_{P2})\\
        \end{bmatrix}\\
    \end{equation}

    \begin{align}
    \label{eqn: omega4}
    \begin{split}
        &\Omega_4 = \begin{bmatrix}
            2 \frac{1}{k}
            \begin{bmatrix} \dHdxpone \left(\xdptwo - \FDuzero \right) \end{bmatrix}^T
             H
             \left(\xdptwo - \FDuzero \right) \\
             \textbf{0} \\
        \end{bmatrix},
    \end{split}
    \end{align}
    and  $\FDuzero = - \Hinv C$. 

\end{thm}

Note that $\dHdxpone$ is a $N \times N \times N$ tensor, so computing $\Omega_4$ requires a matrix-tensor multiplication.
Algorithm \ref{alg: pino aghf} provides an efficient approach to avoid explicitly constructing the tensor $\dHdxpone$ and performing matrix-tensor multiplication directly. 
It also details how to efficiently evaluate the analytical expressions in Theorem \ref{thm: Analytical PDE expression} using spatial vector algebra and rigid body dynamics algorithms.
Section \ref{subsec: computation efficiency} discusses the computational efficiency of these algorithms and the overall AGHF evaluation.

\subsection{Pseudospectral Method for solving the AGHF}
\label{subsec: pseudospectral}

The goal of this section is to describe how to apply a pseudospectral MOL to solve the AGHF as in \eqref{eqn:aghf omega}.
Throughout the remainder of this section, we assume without loss of generality that we have scaled the time domain of the dynamics in (OCP) so that the initial time is $-1$ and the final time is $1$ instead of $0$ and $T$, respectively.
This assumption is made with any loss of generality because one can shift and scale the dynamics in time to satisfy the assumption.
Note, we make this assumption to simplify the presentation of the pseudospectral method that relies on Chebyshev polynomials whose domain is $[-1,1]$.

At a high level, the pseudospectral MOL begins by representing the solution of the AGHF PDE as a linear combination of Chebyshev polynomials of $t \in [-1,1]$ at each value of $s$. 
Taking inspiration from pseudospectral methods, we represent the Chebyshev function by its values at certain discrete points in $t$, which are called the collocation nodes.
Notably, computing the values of the derivatives of the function at these same discrete nodes can be done by applying a matrix called the differentiation matrix. 
As a result, the AGHF PDE at each of the collocation nodes can be written down as a system of ordinary differential equations. 
Once the values of the solution are known at each of the collocation nodes at some final $s$, then one can apply Chebyshev interpolation to construct the steady state solution of the AGHF. 

Recall in Section \ref{sec:intro}, we described how the regular method of lines works by discretizing the PDE in one dimension to generate a set of collocation nodes, and then approximating derivatives at these collocation nodes using finite difference.
The benefits of applying a pseudospectral method of lines as opposed to the regular method of lines, lies in the ability for the pseudospectral method to represent the solution to the PDE as a polynomial, which simultaneously enables it to give high accuracy derivatives at each of the nodes.
In particular, by using a polynomial basis set (i.e., the Chebyshev Polynomials), the number of nodes to achieve a high accuracy solution is significantly less than the number of nodes required by the classical MOL algorithm \cite[Chapter 2, Chapter 4]{boyd2001chebyshev}. 

The remainder of this subsection describes how to perform the transformation from the AGHF PDE to a system of ordinary differential equations. 
To begin, let $p \in \mathbb{N}$ and define the \emph{Chebyshev nodes} as:
\begin{equation}
\label{eqn: chebnodes}
    t_i = -\cos\left(\frac{\pi i}{p}\right),
\end{equation}
for each $i \in \{0,\ldots,p\}$.  
If we fix a particular $p \in \mathbb{N}$ and have the values of a continuously differentiable function at all of the Chebyshev nodes, then we can compute the approximate the values of the derivative of that function at the Chebyshev nodes by using the differentiation matrix, $D:\mathbb{R}^{p+1} \to \mathbb{R}^{p+1}$ \cite[(21.2)]{trefethen2019approximation}.
Note that one that just needs to multiply this matrix by the vector of function values at the Chebyshev nodes to compute the approximate value of the derivative of the function at the Chebyshev nodes.
In fact, the differentiation matrix of size $p+1$ generates the exact derivative at the Chebyshev nodes for polynomials of degree $p$ or less. 
Note that for suitably smooth functions one can compose this differentiation matrix to compute higher order derivatives (i.e., $D^3$ can be applied to compute the third derivative of a function at the Chebyshev nodes).

Next, we transform the AGHF PDE \eqref{eqn:aghf omega} into a system of ODEs.
To do this, fix $p \in \mathbb{N}$ and for each $s$ denote the value of the solution at a particular Chebyshev node $t_i$ as $\psX_i(s) = \X^{T}(t_i,s)$ and let 
\begin{equation}
\label{eqn: vector chebfun}
\psX(s) = \begin{bmatrix} \psX_0(s) \\ \vdots \\ \psX_p(s) \end{bmatrix} \in \mathbb{R}^{(p+1) \times 2N}.
\end{equation}
By using the definition of the differentiation matrix, notice that the $i$th row of $D \psX(s)$ is an approximation of $\frac{\partial {\X}^T}{\partial t}(t_i,s)$.
For convenience let us denote the $i$th row of $D \psX(s)$ as $[D\psX]_i(s)$. 

With these definitions, we can write down the AGHF PDE \eqref{eqn:aghf omega} at each of the nodes as system of ODEs:
\begin{equation}
\label{eqn: vector ps aghf}
\frac{d \psX}{ds}(s) = \begin{bmatrix} \frac{d \psX_0}{ds}(s) \\ \vdots \\ \frac{d \psX_p}{ds}(s) \end{bmatrix} \in \mathbb{R}^{(p+1) \times 2N}
\end{equation}
where
\begin{equation}
\label{eq:ps aghf}
\frac{d \psX_i}{ds}(s) = \Omega \big( \psX^T_i(s),  [D\psX]^T_i(s), [D^2\psX]^T_i(s), k \big)
\end{equation}
for each $i \in \{0,\ldots,p\}$.
For notational convenience, we have abused notation and have not transposed $\frac{d \psX_i}{ds}(s)$ on the left hand side of the previous equation.

This system of ordinary differential equations can be simultaneously solved by using an appropriate differential equation solver. 
Though we do not prove it here, one can show that under certain regularity assumptions regarding the numerical method used to solve the ODEs that the solution computed by the pseudospectral method of lines converges to the true solution of the PDE \cite[Chapter 9 and 12]{boyd2001chebyshev}.

Finally, note for each $p \in \N$, $t_0$ and $t_p$ correspond to $-1$ and $1$, respectively. 
As a result, one can ensure that the initial and final state of the AGHF solution for all $s$ satisfies the boundary conditions by setting $\psX_0(s) = \X_0$ and $\psX_n(s) = \X_f$.
Note this allows us to reduce the number of system of ODEs by $4N$.

The \methodname{} algorithm is summarized in Algorithm \ref{alg: psaghf}.
It first requires one to specify some initial curve $\X_0$ with initial and terminal points as $\x_0$ and $\x_f$ respectively.
Along with the initial curve one must specify the number of pseudospectral nodes $p$, the penalty term $k$ to be used in $G$, and the final $s$ in the domain of the homotopy $s_{max}$.
Algorithm \ref{alg: psaghf} then sets the values of the initial pseudospectral nodes, $\psX(0)$, equal to the initial curve  (Line \ref{alg2:initial}).
Then an ODE solver (e.g., Runge Kutta or Adams Bashforth Method \cite{boyd2001chebyshev}) can be used to simulate solution of the AGHF PDE at each of the collocation nodes  (Line \ref{alg2:solve}).
To evaluate the dynamics within the ODE, one can apply Algorithm \ref{alg: pino aghf}.
Finally, one can extract the control input at each of the collocation nodes by using \eqref{eqn:control extraction} (Line \ref{alg2:extract}). 
Note that one could also perform Chebyshev interpolation to compute the control input for all $t \in [-1,1]$ \cite{trefethen2019approximation}.

\begin{algorithm}[t]
    \caption{$\methodname$}
    \label{alg: psaghf}
    \begin{algorithmic}[1]
        \REQUIRE $\X_0:[0,T] \to \R^{2N}$ s.t. $\X_0(-1) = \x_0$, $\X_0(1) = \x_f$, $p \in \N$, $k$ and $s_{max}$. \label{alg2:require}
        \STATE $\psX^T_i(0) \leftarrow \X_0(t_i)$ for $i \in\{1,\ldots,p-1\}$ \eqref{eqn: chebnodes}. \label{alg2:initial}
        \STATE \textbf{Compute} $\psX(s_{max})$ using an ODE Solver. \label{alg2:solve}
        \STATE \textbf{Compute} $\uu(t_i)$ using $\psX(s_{max})$ and \eqref{eqn:control extraction}. \label{alg2:extract}
    \end{algorithmic}
\end{algorithm}

\subsection{Computing the $\methodname$ Jacobian Partial Derivatives Analytically}
\label{subsec: jacobian}

To solve the system of ODEs in \eqref{eqn: vector ps aghf}, we leverage a differential equation solver that uses an implicit method \cite{pareschi2000implicit}. 
This implicit method requires the derivative of \eqref{eq:ps aghf}.
Most ODE solvers that use implicit methods approximate the derivative numerically.
This can reduce accuracy and increase the number of function evaluations, slowing down the process.
To avoid this and further speed up \methodname{}, we compute and provide the analytical Jacobian of the system of ODEs with respect to 
$\psX (s)$.
Computing this Jacobian, requires one to compute the Jacobian of $\frac{d \psX_i}{ds}(s)$ with respect to $\psX_i$.
We summarize the form of this Jacobian as a function of the first- and second-order derivatives of the rigid body dynamics in Theorem \ref{thm: PDE jacobian}, whose proof can be found in Appendix \ref{sec: AGHF Jacobian Appendix}.
To efficiently compute the required second-order derivatives, we leverage some of the algorithms highlighted in \cite{singh2024second}.
Algorithm \ref{alg: pino aghf jacobian} shows how we rapidly compute all the necessary terms to evaluate the Jacobian.
Once again, for notational convenience we have abused notation and left out the transpose for $\frac{d \psX_i}{ds}(s)$.

\begin{thm}
\label{thm: PDE jacobian}
    Let $\psX_i(s) = \X^{T}(t_i,s)$, $[D\psX]_i(s) = \Xd^T(t_i,s)$ and $[D^2\psX]_i(s) = \Xdd^T(t_i,s)$.
    Then the Jacobian of $\frac{d \psX_i}{ds}(s)$ with respect to $\psX_i (s)$,  $\Jpi(s)$ is given by:
    \begin{equation}
    \label{eqn: jacobian}
    \begin{split}
        & \Jpi = \frac{d \big(\frac{d \psX_i}{ds}(s) \big)}{d\psX_i (s)} = \frac{ d\Omega \big( \psX^T_i(s),  [D\psX]^T_i(s), [D^2\psX]^T_i(s), k \big)}{d\psX_i (s)} \\
        & = \frac{d \Omega}{d \psX_i} \frac{d \psX_i}{d \psX_i} + \frac{d \Omega}{d [D\psX]_i} \frac{d [D\psX]_i}{d \psX_i} + \frac{d \Omega}{d [D^2\psX]_i} \frac{d [D^2\psX]_i}{d \psX_i}
    \end{split}
    \end{equation}
    where 
    \begin{align}
    \label{eqn: d_omega_d_psi}
    \begin{split}
         & \hspace{6.8cm} \frac{d \Omega}{d \psX_i} = \frac{d \Omega_1}{d\X} - (\dOmegatwodx - \dOmegathreedx +\dOmegafourdx) \\
         & = \frac{d \Omega}{d \psX_i} \bigg(H, \Hdot, C, \Cdot, \FDuzero, \dHdxpone, \ddHddxpone, \dHdotdxpone, \dmHinvCdxpone, \dmHinvCdxptwo, \dCdxpone, \dCdxptwo, \dCdotdxpone, \dCdotdxptwo, \ddFDddxpone, \ddFDddxptwo, \ddFDdxponetwo \bigg)
    \end{split}
    \end{align}

    \begin{align}
    \begin{split}
        & \hspace{2.2cm} \frac{d \Omega}{d [D\psX]_i}
        = \frac{d \Omega_1 }{d\Xd} 
        - (\dOmegatwodxd - \dOmegathreedxd +\dOmegafourdxd) \\
        & = \frac{d \Omega}{d [D\psX]_i} \bigg(H, C, \Hdot, \FDuzero, \dHdxpone, \dmHinvCdxpone, \dmHinvCdxptwo, \dCdxpone, \dCdxptwo \bigg)
    \end{split}
    \end{align}

    \begin{align}
    \begin{split}
        \frac{d \Omega}{d [D^2\psX]_i}
        = 2 I_{2N \times 2N}
    \end{split}
    \end{align}

\end{thm}

\subsection{Rapidly evaluating the AGHF \eqref{eqn:aghf omega} and its Jacobian \eqref{eqn: jacobian}}
\label{subsec: computation efficiency}

This section presents several algorithms to rapidly evaluate the AGHF in \eqref{eqn:aghf omega} and its Jacobian \eqref{eqn: jacobian} using spatial vector algebra based rigid body dynamics algorithms and discusses the computational efficiency of these algorithms.

\subsubsection{Rapidly evaluating the AGHF \eqref{eqn:aghf omega}}
\label{subsubsec: computation efficiency AGHF}

To efficiently evaluate the AGHF at each of the collocation nodes, one can use Theorem \ref{thm: Analytical PDE expression}.
Algorithm \ref{alg: pino aghf} shows how to leverage spatial vector algebra and some state-of-the-art dynamics algorithms \cite{pinocchio} to rapidly compute the expressions introduced in Algorithm \ref{alg: pino aghf}.
By using recursive algorithms based on spatial vector algebra to compute the necessary dynamics terms in \eqref{eqn:aghf omega}, this approach provides a substantial speedup over \cite{AGHF_OG}.
For example, algorithms like the Recursive Newton-Euler Algorithm (RNEA) achieve $\OO(N)$ complexity for systems with $N$ bodies, efficiently propagating forces and accelerations throughout the robot's kinematic chain and enabling faster, more scalable evaluation of the dynamics terms. 
This combination of recursive methods allows for a more rapid evaluation of the AGHF’s right-hand side compared to the original approach in \cite{AGHF_OG}, especially for higher dimensional systems.
Table \ref{table: AGHF evaluation times} in Section \ref{sec:experiments}, shows how well this algorithm scales with increasing system dimension compared to \cite{AGHF_OG}.

\begin{algorithm}[t]
    \caption{Leveraging Spatial Vector Algebra to Compute $\Omega$ \eqref{eqn:aghf omega}}
    \label{alg: pino aghf}
    \begin{algorithmic}[1]
        \REQUIRE ${\X}$, ${\dot{\X}}$, $\ddot{\X}$, ${k}$\\
        \STATE $H$, $\Hdot \leftarrow$ \texttt{CRBA\_D}($\X_{P1}$, $\dot{\X}_{P1}$) \vspace{5pt} \label{alg_rhs: CRBA_D}
        \STATE $C \leftarrow$ \texttt{RNEA}($\X_{P1}$, $\X_{P2}$, \textbf{0}) \vspace{5pt} \label{alg_rhs: RNEA}
        \STATE $\dCdxpone$, $\dCdxptwo \leftarrow$ \texttt{RNEA\_D}($\X_{P1}$, $\X_{P2}$, \textbf{0}) \vspace{5pt} \label{alg_rhs: RNEA_D}
        \STATE $\Cdot \leftarrow \dCdxpone \xdpone + \dCdxptwo \xdptwo$ \vspace{5pt} \label{alg_rhs: C_dot}
        \STATE $\dmHinvCdxpone$, $\dmHinvCdxptwo$, $\Hinv$, $\FDuzero \leftarrow$ \texttt{ABA\_D}($\xpone$, $\xptwo$, \textbf{0}) \vspace{5pt} \label{alg_rhs: ABA_D}
        \STATE \textbf{Compute} $\Omega_1$ \eqref{eqn: ginv_dt_dl_dxd}, $\Omega_2$ \eqref{eqn: omega2} and $\Omega_3$ \eqref{eqn: omega3} \vspace{5pt} \label{alg_rhs: omegas}
        \STATE model\_gravity $\leftarrow 0$ \vspace{5pt} \label{alg_rhs: model_grav}
        \STATE $\dHdxpone(\xdptwo - \FDuzero) \leftarrow$ \texttt{RNEA\_D}($\xpone$, \textbf{0},  $\xdptwo - \FDuzero$) \vspace{5pt} \label{alg_rhs: dHdxpone}
        \STATE $\omega_4 \leftarrow 2 \frac{1}{k} \begin{bmatrix} \dHdxpone(\xdptwo - \FDuzero) \vspace{1.1pt} \end{bmatrix}^T H (\xdptwo - \FDuzero)$ \label{alg_rhs: omega4}
        \STATE $\Omega_4 \leftarrow \begin{bmatrix} \omega_4^T & \textbf{0}^T \end{bmatrix}^T$ \vspace{5pt} \label{alg_rhs: Omega4}
        \STATE \textbf{Compute} $\Omega$ \eqref{eqn:aghf omega} \label{alg_rhs: Omega}
    \end{algorithmic}
\end{algorithm}

In Algorithm \ref{alg: pino aghf}, we begin by computing $H$ and $\Hdot$ using a modified version of the Composite Rigid Body Algorithm (\texttt{CRBA}) (Line \ref{alg_rhs: CRBA_D}).
This modified version leverages the chain rule to compute the time derivatives of the various spatial quantities as we traverse the rigid body tree, yielding the time derivative of the mass matrix $(\Hdot)$.
This modified version, we call \texttt{CRBA\_D}, is a worst-case $\OO(N^3)$ algorithm.
We then use the Recursive Newton Euler Algorithm (\texttt{RNEA}) to rapidly compute $C$ $(\OO(N))$ (Line \ref{alg_rhs: RNEA}) and an extended version (\texttt{RNEA\_D}) that computes \texttt{RNEA}'s derivatives with respect to $\q$, $\qd$ and $\qdd$ to compute the derivatives of $C$ with respect to to $\xpone$ and $\xptwo$ $(\OO(N^2)$ worst-case) (Line \ref{alg_rhs: RNEA_D}).
These are all used to compute $\dot{C}$ (Line \ref{alg_rhs: C_dot}).

Next, we use the algorithm introduced in \cite{carpentierderivs} to compute multiple partial derivatives of the Forward Dynamics when $u=\textbf{0}$ (Line \ref{alg_rhs: ABA_D}) leveraging the Articulated Body Algorithm (\texttt{ABA}).
We denote this as \texttt{ABA\_D}.
This is a worst-case $\OO(N^3)$ algorithm \cite{Singh_2022}.
Utilizing the earlier results, we compute $\Omega_1$, $\Omega_2$ and $\Omega_3$ (Line \ref{alg_rhs: omegas}).
Next, we compute $\dHdxpone(\xdptwo - \FDuzero)$ efficiently by setting the gravity term used by our dynamics model ($\text{model\_gravity}$) to zero (Line \ref{alg_rhs: model_grav}) and using \texttt{RNEA\_D} with zero velocity and setting the acceleration to $(\xdptwo - \FDuzero)$, which avoids explicitly computing the tensor $\dHdxpone$ (Line \ref{alg_rhs: dHdxpone}).
Lastly, we perform a matrix-vector multiplication to compute $\omega_4$ (Line \ref{alg_rhs: omega4}) and stack the vector to obtain $\Omega_4$ (Line \ref{alg_rhs: Omega4}). 
With all the terms computed we can compute $\Omega$ using \eqref{eqn:aghf omega} (Line \ref{alg_rhs: Omega}).

\begin{figure}[h]
    \centering
    \includegraphics[width=0.8\columnwidth]{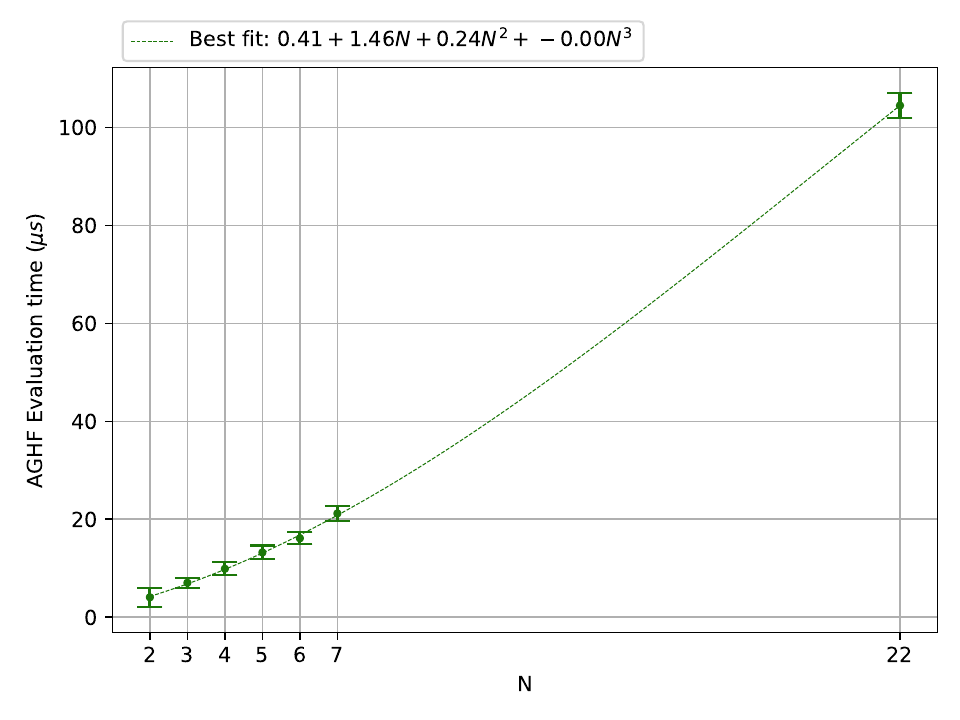}
    \caption{
    Scaling trend of the mean evaluation times (in {$\mu$s}) of the right-hand side of the AGHF using Algorithm \ref{alg: pino aghf} as the number of bodies (N) increases from 2 to 22. The systems with $N$ between 2 and 6 are penduli systems (i.e., 2-link pendulum, 3-link pendulum, etc. ). The $N = 7$ system is the Kinova Gen3 and the $N = 22$ system is the pinned Digit V3 biped. Each robot's AGHF RHS was evaluated at a 1000 random robot configurations.
    Notice the polynomial line of best fit lacks a significant $N^3$ term, suggesting that the right-hand side computation time scales approximately quadratically with $N$ in practice.
    }
    \label{fig:AGHF_evalution_time}
    \vspace*{0cm}
\end{figure}

Combining these operations with the $\OO(N^3)$ matrix-matrix multiplications needed to compute \eqref{eqn:aghf omega}, results in a worst-case $\OO(N^3)$ algorithm for computing the AGHF right-hand side.
Figure \ref{fig:AGHF_evalution_time} shows the mean and standard deviation of the computation time of \eqref{eqn:aghf omega} as the number of bodies ($N$) of the system increases from 2 (Double Pendulum) to 22 (Digit).
While the worst-case computational complexity is $\OO(N^3)$, the results from Figure \ref{fig:AGHF_evalution_time} indicate that the algorithm scales more efficiently in practice.
Specifically, the polynomial line of best fit lacks a significant $N^3$ term, suggesting that the computational time scales approximately quadratically with the number of bodies $N$ within the observed range.

\subsubsection{Rapidly Evaluating the AGHF Jacobian \eqref{eqn: jacobian}}
\label{subsubsec: computation efficiency jacobian}

\begin{figure}[t]
    \centering
    \includegraphics[width=0.8\columnwidth]{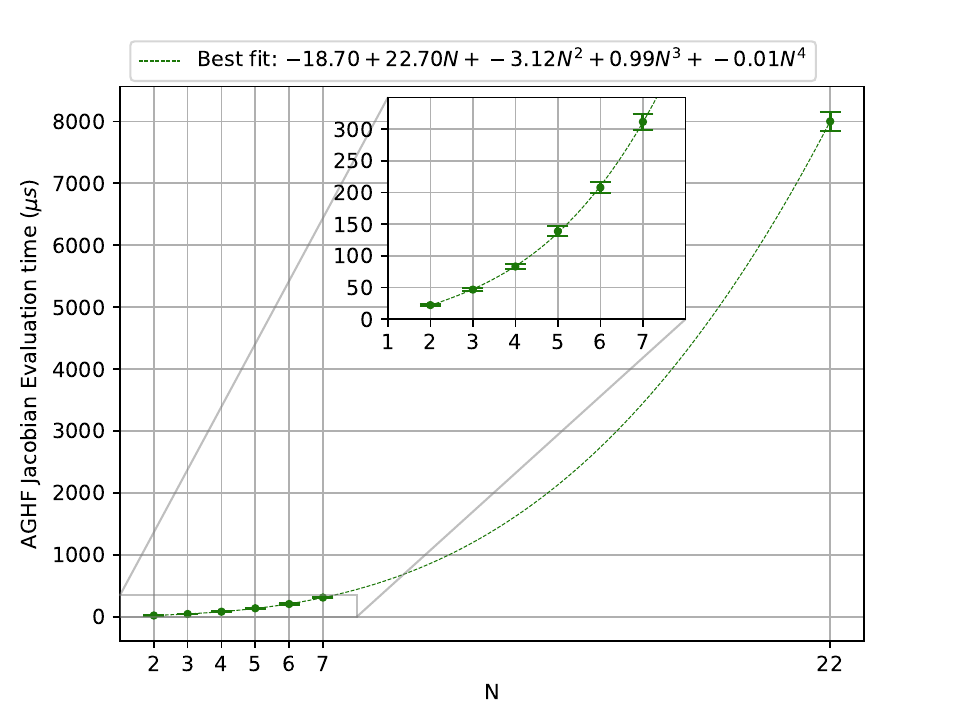}
    \caption{
    Scaling trend of the mean evaluation times (in {$\mu$s}) of the AGHF Jacobian using Algorithm \ref{alg: pino aghf jacobian} as the number of bodies (N) increases from 2 to 22. The systems with $N$ between 2 and 6 are penduli systems (i.e., 2-link pendulum, 3-link pendulum, etc. ). The $N = 7$ system is the Kinova Gen3 and the $N = 22$ system is the pinned Digit V3 biped. Each robot's AGHF Jacobian was evaluated at a 1000 random robot configurations.
    Notice the polynomial line of best fit lacks a significant $N^4$ term, suggesting that the Jacobian computation time scales approximately cubically with $N$ in practice.
    }
    \label{fig:AGHF_Jacobian_evalution_time}
    \vspace*{-0.5cm}
\end{figure}

To efficiently evaluate the AGHF Jacobian at each of the collocation nodes, one can use Theorem \ref{thm: PDE jacobian}.
Algorithm \ref{alg: pino aghf jacobian} shows how to leverage spatial vector algebra and state-of-the-art dynamics algorithms \cite{pinocchio} to rapidly compute this Jacobian in \eqref{eqn: jacobian}.
We begin by first computing $H$, $\Hdot$, $\dHdxpone$ and $\ddHddxpone$ (Line \ref{alg_jac: CRBA_2D}) using a modified \texttt{CRBA} algorithm where we compute $\Hdot$ and the first and second derivatives of $H$ with respect to $\xpone$ using the chain rule.
This modified version, we call \texttt{CRBA\_2D}, is a worst-case $\OO(N^4)$ algorithm.
Similar to Section \ref{subsubsec: computation efficiency AGHF}, we use \texttt{ABA\_D} (worst-case $\OO(N^3)$\cite{Singh_2022}) to compute the partial derivatives of the Forward Dynamics (Line \ref{alg_jac: ABA_D}).
We then use \texttt{RNEA} to compute $C$ $(\OO(N))$ (Line \ref{alg_jac: RNEA}) and \texttt{RNEA\_D} to compute the derivatives of $C$ (Line \ref{alg_jac: RNEA_D}) once more ($(\OO(N^2)$ worst-case).
Next we use these derivatives to compute $\dot{C}$ (Line \ref{alg_jac: C_dot}).
Next, the function $\texttt{get\_Hdot\_D}$ computes $\dHdotdxpone$ by applying the chain rule to $\ddHddxpone$ and $\xdpone$ (Line \ref{alg_jac: Hdot_D}).
We then use \texttt{RNEA\_2D}, which computes the second derivatives of the Inverse Dynamics (ID), with the acceleration passed in as zero to get the second derivatives of $C$ (Line \ref{alg_jac: C_partials}).
Next, \texttt{RNEA\_2D} with the acceleration set to $\FDuzero$ (which is the acceleration of the system, with $u=0$) allows us to compute the derivatives of the inverse dynamics (Line \ref{alg_jac: RNEA_2D}).
The inverse dynamics derivatives computed using the acceleration from $\FDuzero$ are needed to to rapidly compute the second derivatives of $\FDuzero$  using the algorithm proposed in \cite{singh2024second}, \texttt{ABA\_2D} (Line \ref{alg_jac: ABA_2D}).
The function $\texttt{get\_Cdot\_D}$, similar to $\texttt{get\_Hdot\_D}$, computes the partial derivatives of $\dot{C}$ using the chain rule with the second derivatives of $C$ and $\xdpone$ and $\xdptwo$ (Line \ref{alg_jac: C_dot_D}).
With all these terms computed we then evaluate $\Jpi(s)$ \eqref{eqn: jacobian} (Line \ref{alg_jac: J}).
Overall, combining these operations with the $\OO(N^4)$ tensor-matrix multiplications needed to compute \eqref{eqn: jacobian}, results in an $\OO(N^4)$ algorithm for computing $\Jpi(s)$.
Figure \ref{fig:AGHF_Jacobian_evalution_time} shows the mean and standard deviation of the computation time of $\Jpi(s)$ as the number of bodies ($N$) of the system increases from 2 (Double Pendulum) to 22 (Digit).
While the worst-case computational complexity is $\OO(N^4)$, the results from Figure  \ref{fig:AGHF_Jacobian_evalution_time} indicate that the algorithm scales more efficiently in practice.
Specifically, the polynomial line of best fit lacks a significant $N^4$ term, suggesting that the Jacobian computational time scales approximately cubically with the number of bodies $N$ within the observed range.

\begin{algorithm}[t]
    \caption{Leveraging Spatial Vector Algebra to Compute $\Jpi(s)$ \eqref{eqn: jacobian}}
    \label{alg: pino aghf jacobian}
    \begin{algorithmic}[1]
        \REQUIRE ${\X}$, ${\dot{\X}}$, $\ddot{\X}$, ${k}$\\
        \STATE $H$, $\Hdot$, $\dHdxpone$, $\ddHddxpone \leftarrow$ \texttt{CRBA\_2D}($\X_{P1}$, $\dot{\X}_{P1}$) \vspace{5pt} \label{alg_jac: CRBA_2D}
         \STATE $\dmHinvCdxpone$, $\dmHinvCdxptwo$, $\Hinv$, $\FDuzero \leftarrow$ \texttt{ABA\_D}($\xpone$, $\xptwo$, \textbf{0}) \vspace{5pt} \label{alg_jac: ABA_D}
         \STATE $C \leftarrow$ \texttt{RNEA}($\X_{P1}$, $\X_{P2}$, \textbf{0}) \vspace{5pt} \label{alg_jac: RNEA}
        \STATE $\dCdxpone$, $\dCdxptwo \leftarrow$ \texttt{RNEA\_D}($\X_{P1}$, $\X_{P2}$, \textbf{0}) \vspace{5pt} \label{alg_jac: RNEA_D}
        \STATE $\Cdot \leftarrow \dCdxpone \xdpone + \dCdxptwo \xdptwo$ \vspace{5pt} \label{alg_jac: C_dot}
        \STATE $\dHdotdxpone \leftarrow$ \texttt{get\_Hdot\_D}($\xdpone$, $\ddHddxpone$)  \vspace{5pt} \label{alg_jac: Hdot_D}  
        \STATE $\ddCddxpone$, $\ddCddxptwo$, $\ddCdxponetwo \leftarrow$ \texttt{RNEA\_2D}($\X_{P1}$, $\X_{P2}$, \textbf{0}) \vspace{5pt} \label{alg_jac: C_partials}
        \STATE $\ddIDddxpone$, $\ddIDddxptwo$, $\ddIDdxponetwo \leftarrow$ \texttt{RNEA\_2D}($\X_{P1}$, $\X_{P2}$, $\FDuzero$) \vspace{5pt} \label{alg_jac: RNEA_2D}
        \STATE $\ddFDddxpone$, $\ddFDddxptwo$, $\ddFDdxponetwo \leftarrow$ \texttt{ABA\_2D}($\dmHinvCdxpone$, $\dmHinvCdxptwo$, $\Hinv$, $\dHdxpone$, $\ddIDddxpone$, $\ddIDddxptwo$, $\ddIDdxponetwo$) \vspace{5pt} \label{alg_jac: ABA_2D}
        \STATE $\dCdotdxpone$, $\dCdotdxptwo$ $\leftarrow$ \texttt{get\_Cdot\_D}($\xdpone$, $\xdptwo$, $\ddCddxpone$, $\ddCddxptwo$, $\ddCdxponetwo$, $\ddCdxptwoone)$ \vspace{5pt} \label{alg_jac: C_dot_D}
        \STATE \textbf{Compute} $\Jpi(s)$ \eqref{eqn: jacobian} \vspace{5pt} \label{alg_jac: J}
    \end{algorithmic}
\end{algorithm}

\section{Incorporating constraints into the AGHF}
\label{sec: AGHF constraints}

This section explains how constraints are incorporated into the AGHF by adding them to the Lagrangian. 
The constraint terms are designed so that any violations increase the magnitude of the action functional. 
As we minimize the action functional, \methodname{} naturally converges to solutions that satisfy the constraints.
Below we discuss the form and properties of the added constraint term and show how it augments Lagrangian and the AGHF PDE.
Note that similar to Theorems \ref{thm: Analytical PDE expression} and \ref{thm: PDE jacobian}, in the subsequent subsections we denote $\X(t,s)$ by $\X$.

\subsection{Constraint Lagrangian}
\label{subsec: AGHF constraint}

We incorporate constraints into the AGHF by using a penalty term in the Lagrangian in a similar fashion to \cite{fan2019midair}.
This penalizes the PDE when it evolves towards undesirable states.
By adding the penalty term, our original Lagrangian from \eqref{eqn:lagrangian} is augmented in the following way:
\begin{defn}\label{defn: constraint lagrangian}
Let $k_{cons}$ be some large positive integer that penalizes constraint violation, and let $g_j(\X)$ be the $j$-th inequality constraint evaluated at $\X$.
Finally, let $L$ be the Lagrangian from \eqref{eqn:lagrangian} with additional terms to enforce the constraints for all $j \in \J$.
$L_{cons}$ is given by:
\begin{equation}
\label{eq: constraint lagrangian}
    L_{cons}(\X,\dot{\X}, g_j(\X))  = L(\X,\dot{\X})  + \sum_{j \in \J} b(g_j(\X))
\end{equation}
where
\begin{equation}
\label{eq: constraint penalty}
    b(g_j(\X)) = k_{cons} \cdot (g_j(\X))^2 \cdot S(g_j(\X)),
\end{equation}
where $S:\R \to \R$ is defined as follows:
\begin{enumerate}
    \item $S:\R \to \R$ is a positive, differentiable function, 
    \item $S(g_j(\X)) = 0$ when $g_j(\X) \leq 0$ and
    \item $S(g_j(\X)) = 1$ when $g_j(\X) > 0$.
\end{enumerate}
\end{defn}
An example of an $S$ satisfying this definition is:
\begin{equation}
\label{eq: barrier lagrangian}
    S(g_j(\X)) = \frac{1}{2} + \frac{1}{2} \tanh(\ccons \cdot g_j(\X))
\end{equation}
where $\ccons$ is a hyper-parameter that determines how fast $S(g_j(\X))$ transitions from 0 to 1, once the constraint is violated. 
We introduce the form of the AGHF with $L_{cons}$ in the following definition:
\begin{defn}
Let $L_{cons}$ be used as the Lagrangian to construct the AGHF PDE.
Then the constrained AGHF PDE is given by:
\label{eqn: constraint aghf}
 \begin{equation}
    \dxds = \Ginv(\X) \bigg(\ddt \dldxd - \dldx - \sum_{j \in \J} \dbdxg \bigg)
 \end{equation}
\end{defn}

The proof of convergence of the constrained AGHF is given in the proof of \cite[Lemma 4.1]{fan2021thesis}.
\section{Experiments and Results}
\label{sec:experiments}
This section evaluates the speed and performance of $\methodname$ on a variety of scenarios and robot platforms and compares $\methodname$ against the state of the art trajectory optimization algorithms.
We begin by explaining the experimental setup. 
Next, we discuss various implementation details of the comparison methods and conclude the section by summarizing the different evaluations.

\subsection{Experimental Setup}
\label{subsec:setup}

\subsubsection{Robot Platforms and Comparison Methods}
We perform evaluations on the following robot platforms: the 1 to 5 link pendulums in two dimensions, the Kinova Gen3 7DOF arm, and a pinned version of Digit V3 by Agility Robotics with the closed kinematic chains removed. 
Note we have removed the kinematic chain because a version of the RNEA algorithm that incorporates such a chain has not been implemented within the spatial vector algebra and rigid body dynamics package that we utilize -- Pinnochio \cite{pinocchio}.
We compare the performance of $\methodname$ to the original AGHF \cite{AGHF_OG}, Crocoddyl \cite{crocoddyl2020}, and Aligator \cite{aligator}.
Note we refer to Crocoddyl and Aligator as the trajectory optimization methods as in contrast to the $\methodname$ and the original AGHF approach they rely on optimization to synthesize a trajectory.
We choose to compare to these methods in particular because these recent DDP methods have been shown to outperform the collocation-based methods \cite{wensing_ocp_summary} in terms of computation time and the other indirect and dynamic programming methods mentioned in Section \ref{sec:intro} do not scale well to high-dimensional robots.
Details on the scenarios used for the comparisons ($\x_0$, $\x_f$, obstacles description etc.) are available online at \url{https://github.com/roahmlab/PHLAME_comparisons}.

\subsubsection{Determining Success or Failure}
\label{subsubsec: success_failure}
Evaluating whether a solution generated by a numerical method is satisfactory is non-trivial.
In particular, each of these numerical methods compute an open loop input, $\uu^*:[0,T] \to \R^m$ and associated trajectory of the system, $\X^*:[0,T] \to \R^{2N}$; however, it is unclear whether these solutions can actually be successfully followed by the robot. 
For instance, when one forward integrates the system using the control input synthesized by one of the evaluated algorithms, one may find that it violates the constraints. 
One could check constraint satisfaction by applying this open-loop control directly into the high dimensional robotic system and checking whether constraints are satisfied; however, this is also impractical because small numerical integration errors can compound significantly over time, causing the actual trajectory to deviate substantially from the planned path. 
As a result, to fairly assess whether constraints are satisfied, we instead integrate forward using the following feedback controller:
\begin{equation}
    \Ufb(t) = \uu^*(t) +  k_p (\X^*_{P1}(t) - \q(t)) + k_v (\X^*_{P2}(t) - \qd(t)),
\label{eq: feed forward control}
\end{equation}
where $k_p$ and $k_v$ are parameters that we fix in each experiment and $\q$ and $\qd$ correspond to the position and velocity of the robot whose dynamics we are integrating forward using \eqref{eq: feed forward control}.
Note that when we report the control effort for a particular experiment and algorithm, we compute the $2$-norm of $\Ufb$.

In the case of an experiment without obstacles, a solution obtained by a method is successful if the error between the final state of the forward integrated solution and $\x_f$ is less than some $\epsilon$ in the infinity norm.
Note $\epsilon$ is a constant threshold that is set to $0.05$ for all experiments.
In the case of experiments with obstacles, a trial is successful if the previous condition is met and additionally, the joint position of the forward integrated solution at a time resolution of $\delta t = 10^{-2}$s is not inside any of the obstacles at each joint frame.

\subsubsection{Solver Setup}

All experiments were run on a Ubuntu 22.04 machine with an Intel Xeon Platinum 8170M @ 208x 3.7GHz CPU. 
Each of the evaluated numerical methods requires an initial and final desired state and an initial guess for the initial trajectory $\X_0$ along with the selection of solver specific parameters.
Note the trajectory optimization methods also require an initial control input $\uu_0:[0,T] \to \R^{m}$.
Throughout this section, the term experiment refers to a tuple of experimental parameters consisting of the robotic platform, $\x_0$, $\x_f$, and a set of obstacles (if present).

To perform a fair comparison, we ran a grid search over the parameter space to obtain the best possible solver parameters. 
Each grid search was ran in parallel with a timeout per parameter set.
The timeout for the systems with $\leq5$ DOF was 3 minutes and for the rest, 5 minutes.
Then, we re-ran the experiments sequentially with the best solver specific parameters per method to obtain timings.
The best solver specific parameter is defined as the one that yields the lowest solve time while producing a success for the experiment as defined above.

We next describe the parameters that make up the grid search for each method. 
Note that these parameters and how they are varied are summarized in Apppendix \ref{subsec: results_appendix}.

As part of the grid search for Crocoddyl and Aligator, we considered different types of initial guesses to pass to the optimizer based on the examples given in their publicly available code.
For Aligator, we considered the following initial guesses: (1) ``Zeros,'' which corresponds to an initial guess that is zero for all time for $\X_0$ and $\uu_0$; (2) ``Line and RNEA,'' which corresponds to an initial guess where $\X_0$ is a line connecting $\x_0$ and $\x_f$ and $\uu_0$ is equal to applying RNEA using $\X_0$ and $\Xd_0$ (here,  $\Xd_0$ is obtained by fitting a chebyshev polynomial to $\X_0$ and taking it's derivative); (3) ``Rollout and Zero,'' which corresponds to an initial guess where $\uu_0$ is zero and $\X_0$ is the result of forward simulating the dynamical system using zero input; and (4) ``Rollout and Constant,'' which corresponds to an initial guess for $\uu_0$ that is constant and is equal to applying RNEA to the initial $\x_0$ while assuming that $\qdd(0)=0$ and $\X_0$ is equal to forward simulating the system using that control input. 
For Crocoddyl, we considered the following initial guesses: ``Zeros'' and ``Line and RNEA,'' as defined in the Aligator case, and ``Constants,'' which corresponds to setting $\X(t) = \x_0$ for all $t$ and $\uu_0$ equal to a constant that is equal to applying RNEA to the initial $\x_0$ while assuming that $\qdd(0)=0$ and $\X_0$.
For \methodname{}, we only considered the initial guess ``Line`, which corresponds to an initial guess where $\X_0$ is a line connecting $\x_0$ and $\x_f$. Note that for our method we do not need to specify an initial guess $\uu_0$.

Aligator and Croccodyl have several parameters that are specific to their implementation. 
First, each method relies upon discretizing time and allow a user to specify a time discretization, $\delta_t$.
Second, each method allows one to include a running cost. 
We choose this running cost to be the $2$-norm of the input added to the $2$-norm of the state with weights $w_u$ and $w_x$, respectively. 
Third, each method allows one to include a terminal cost with weight $w_{xf}$.
We choose this to be the $2$-norm of the difference of the final state from $\x_f$. 
For Aligator, we also consider the parameters $\mu_{init}$ and $\epsilon_{tol}$, where the former corresponds to the initial value of the augmented Lagrangian penalty parameter and the latter to the solver tolerance.
Lastly, for Aligator we also add an equality constraint that enforces the desired final state $\x_f$ and inequality constraints that enforce that the joint frames are not inside any of the sphere obstacles.

As for \methodname{}. 
First, $p$ is the degree of the polynomial that represents the solution. 
Second, $k$ is a penalty that ensures dynamic feasibility and was first introduced in \eqref{eqn: definition of K}.
Third, $s_{max}$ corresponds to the maximum "time" that the PDE has to evolve.
Fourth, only for the experiments that have obstacles (inequality constraints) we also consider the parameters $k_{cons}$, introduced in \eqref{eq: constraint penalty} and $\ccons$, introduced in \eqref{eq: barrier lagrangian} which control the weight of the constraint satisfaction and the sharpness of the activation function respectively.

In summary, for Crocoddyl in each experiment we perform a grid search of Initial guess, $\delta_t, w_{u}, w_{x},$ and $w_{xf}$.
For Aligator we search over the Initial guess, $\delta_t, w_{u}, w_{x}, w_{xf}, \mu_{init}$  and $\epsilon_{tol}$.
For unconstrained \methodname{}, $p$, $s_{max}$ and $k$ and for constrained \methodname{} $p$, $s_{max}$, $k$, $\ccons$ and $k_{cons}$.

\subsection{\methodname{} AGHF Evaluation vs Original AGHF \cite{AGHF_OG}}

This section compares evaluating the right hand side of the AGHF PDE using the original Matlab based implementation \cite{AGHF_OG} and using Algorithm \ref{alg: pino aghf}.
We compare the evaluation time as the number of bodies increases from 2 to 22.
The systems with $N$ between 2 and 6 are penduli systems (i.e., 2-link pendulum, 3-link pendulum, etc.).
The $N = 7$ system is the Kinova Gen3 and the $N = 22$ system is the pinned Digit V3 biped.
We evaluate the AGHF at a 1000 random configurations for all of these systems.
Table \ref{table: AGHF evaluation times} shows the evaluation times of the AGHF using \methodname{} and the original AGHF \cite{AGHF_OG}. 

\begin{table}[t]
    \centering
    \begin{tabular}{|c||c|c|}
    \hline 
    \multirow{2}{*}{\textbf{Number of Bodies} (\boldsymbol{$N$})} & \multicolumn{2}{c|}{\textbf{Evaluation Time [$\boldsymbol{\mu}$s]}} \\ \cline{2-3}
     & \textbf{\methodname} & \textbf{AGHF \cite{AGHF_OG}} \\ \hline
    2 & 4.071 $\pm$ 1.963 & 46.819 $\pm$ 11.559 \\ \hline
    3  & 7.029 $\pm$ 1.035 & 593.054 $\pm$ 286.744  \\\hline
    4 & 9.895 $\pm$ 1.297 & 44582.9 $\pm$ 7963.98 \\\hline
    5 & 13.234 $\pm$ 1.411 & 359506 $\pm$ 36460.7 \\\hline
    6 & 16.122 $\pm$ 1.190 & DNF \\\hline
    7 & 21.195 $\pm$ 1.580 & DNF \\\hline
    22 & 104.507 $\pm$ 2.516 & DNF \\\hline
    \end{tabular}
    \caption{
    A table showing mean evaluation times (in {$\mu$s}) of the right-hand side of the AGHF using \methodname{} and AGHF \cite{AGHF_OG} as the number of bodies (N) increases from 2 to 22.
    For $N \leq 5 $ the results correspond to the 1-5 link pendulum, for $N=7$ to the Kinova Gen3 Arm and for $N=22$ for the Pinned Digit biped robot. For each method each robot's AGHF RHS was evaluated at a 1000 random robot configurations each. For $N > 5$ the MATLAB method used in \cite{AGHF_OG} was unable to symbolically generate the AGHF right-hand side without running out of memory. We denote the results where \cite{AGHF_OG} was unable to generate the right-hand side and find a solution as DNF (Did Not Find). Regardless, we see that \methodname{}'s right-hand side evaluation time is much faster and scales better than the original AGHF \cite{AGHF_OG}. }
    \label{table: AGHF evaluation times}
\end{table}

The results demonstrate the significant speedup achieved by Algorithm \ref{alg: pino aghf} in evaluating the AGHF. 
This, combined with the substantial reduction in the number of nodes needed to accurately solve the AGHF, allows \methodname{} to efficiently scale to high-dimensional systems while preserving fast solution times.
The next section discusses how \methodname{}'s overall solve time scales with increasing number of bodies ($N$).

\begin{figure}[t]
    \centering
    \includegraphics[width=0.8\columnwidth]{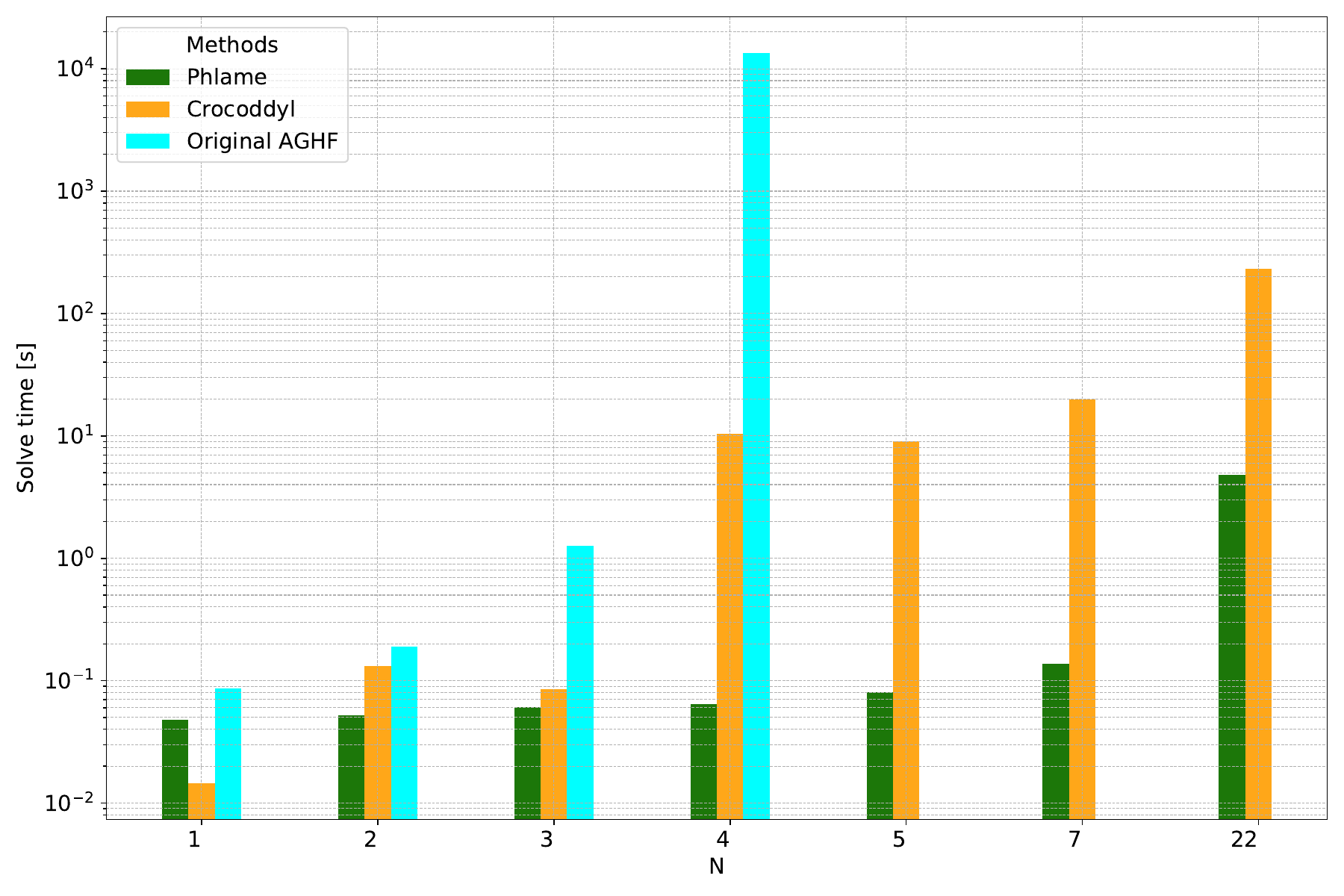}
    
    \caption{
    A bar plot comparing the mean solve times for three different trajectory optimization algorithms: \methodname{}, AGHF \cite{AGHF_OG} and Crocoddyl \cite{crocoddyl2020}. For $N \leq 5 $ the results correspond to the 1-5 link pendulum, for $N=7$ to the Kinova Gen3 Arm and for $N=22$ for the Pinned Digit biped robot. Each experiment was run ten times using the best solver parameter set. For $N>=5$ we do not show results for the Original AGHF implementation as no results were obtained in a reasonable amount of time. Overall, we see that \methodname{} shows better empirical scalability and solve time than the other methods.
    The values of $\x_0$ and $\x_f$ are provided in \url{https://github.com/roahmlab/PHLAME_comparisons}.
    }
    \label{fig:scalability-t-solve-no-obstacles}
    \vspace*{-0.5cm}
\end{figure}

\subsection{Scalability with Increasing State Dimension in Unconstrained Systems}
This section compares the methods Crocoddyl, $\methodname$ and AGHF \cite{AGHF_OG} to solve a fixed time swing up problem for a 1-, 2-, 3-, 4-, and 5-link pendulum model, a fixed time and final state specified trajectory optimization for the Kinova arm, and a fixed time and final state trajectory optimization having the pinned Digit execute a step.
As described in Section \ref{subsec:setup}, we run a grid search in parallel to obtain the best parameter sets for \methodname{} and Crocoddyl per problem; the values of the grid search for Crocoddyl without obstacles can be seen in Table \ref{table: Crocoddyl grid without obstacles}. 
The parameters $w_u, w_{xf}, w_x, dt$ and $\text{initial guess}$ correspond to solver specific parameters.
For \methodname{}, the explored parameter grids for the penduli, Kinova and Digit are in Tables \ref{table: aghf grid pendulum swingup}, \ref{table: aghf grid Kinova without obstacles}, and \ref{table: aghf grid Kinova without obstacles} respectively.
For AGHF, we use the same parameters as for \methodname{}, but with $100$ nodes to discretize the trajectory as that was the minimum we could use to yield a good solution.

Figure \ref{fig:scalability-t-solve-no-obstacles} compares solve time with the best parameter set using the described feedback controller with gains $k_p=100, k_v=100$ for Digit and $k_p=10, k_v=10$ for all other robotic platforms.
The results demonstrate how \methodname{} is able to generate solutions faster than these state-of-the-art methods.
As the system dimension increases, \methodname{}  has a much lower solve time than the comparison methods.

\subsection{Trajectory Optimization for the Kinova Arm without Obstacles}
\label{subsec: kinova no obstacles}

\begin{figure}[h]
    \centering
    \begin{subfigure}[b]{0.49\columnwidth}
        \centering
        \includegraphics[width=\columnwidth]{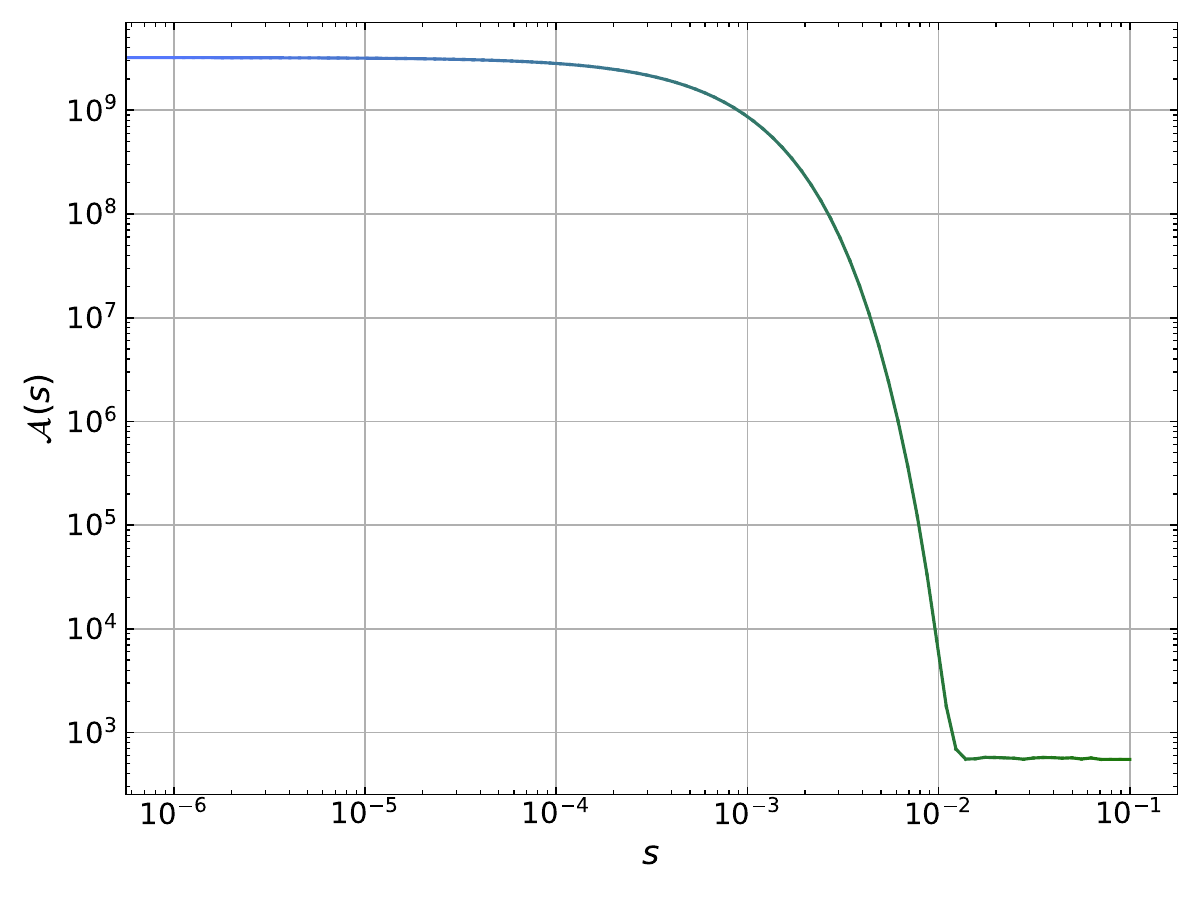}
        \caption{Evolution of the action functional ($\A(s)$) as the \methodname{} evolves the trajectory along $s$ for one of the kinova no obstacles scenarios.}
    \end{subfigure}
    \hfill
    \begin{subfigure}[b]{0.49\columnwidth}
        \centering
        \includegraphics[width=\columnwidth]{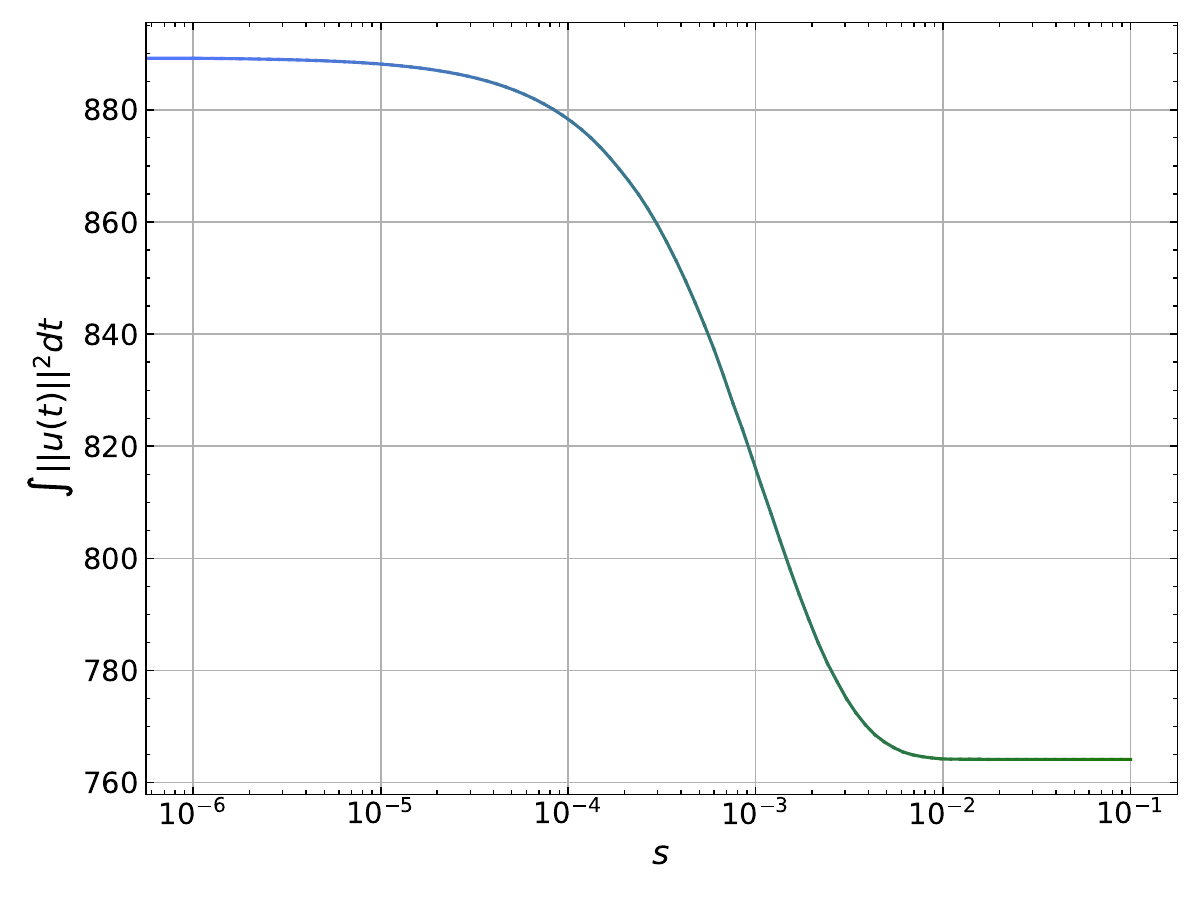}
        \caption{Evolution of $\int_0^T || u(t)||^2 dt $ as the \methodname{} evolves the trajectory  along $s$ for one of the kinova no obstacles scenarios.}
    \end{subfigure}
    \caption{Side-by-side figures showing the convergence of action functional and $\int_0^T || u(t)||^2 dt $ for one of the trajectories found in the kinova no obtacle trajectory optimization problems. The lines transitions from \textcolor{init_traj_blue}{\textbf{dark blue}} to \textcolor{final_traj_green}{\textbf{dark green}} as the trajectory converges to an extremal solution of \eqref{eqn:AGHF}}
    \label{fig:combined kinova no obstacles evolution action and u2}
\end{figure}

We consider ten different scenarios with distinct $\x_0$, $\x_f$.
We select five out of the ten scenarios to perform grid search over to find the best solver specific parameters.
Tables \ref{table: Crocoddyl grid without obstacles} and \ref{table: aghf grid Kinova without obstacles} summarize the grids for Crocoddyl and \methodname{}, respectively.
Then, for each solver method, we find the single solver parameter set that yields the most successes across the five scenarios and the lowest time to solve when using a feedback controller with gains of $k_p=k_v=10$.
The best parameter set for \methodname{} was $s_{max}=0.1, k=10^9, p=9$.
The best parameter set for Crocoddyl was $w_x=0.0001, w_u=0.0001, \delta t=0.0001, w_{xf}=1,$ and $\text{Initial Condition}=\text{``Zeros''}$.
We then run all the scenarios sequentially ten times with the best parameter set and apply the aforementioned feedback controller.

The results are shown in Table \ref{table: comparison kinova no obs}.
For all the scenarios, both methods had a 100\% success rate and were able to generate dynamically feasible trajectories. \methodname{} was faster than Crocoddyl; however, Crocoddyl's trajectory had a smaller controller input. 
Note that \methodname{}'s control input still satisfied the actuation limits of the Kinova arm.

\begin{table}[t]
    \centering
    \begin{tabular}{ |c || c | c | c | c | c| }
    \hline
    \textbf{Method Name} & \textbf{Success Rate} & \boldsymbol{$\int_0^T || \Ufb(t)||^2 dt$} & \textbf{Solve Time [s]} \\
    \hline
    \methodname{} & 10/10 & 430.5 ± 204.8 & 0.2102 ± 0.02454 \\
    \hline
    Crocoddyl & 10/10 & 120.7 ± 50.98& 213.8 ± 82.1\\
    \hline
    \end{tabular}
    \caption{A table summarizing the success rate, norm of the control effort, and solve time in seconds  for Kinova Gen3 trajectory optimization experiment with no obstacles}
    \label{table: comparison kinova no obs}
\end{table}

Figure \ref{fig:combined kinova no obstacles evolution action and u2} illustrates the evolution of the Action Functional and the $\int_0^T || u(t)||^2 dt $ for one of the kinova arm trajectory optimization problems. 
Note that the action functional decreases significantly, ensuring the trajectory's dynamic feasibility while reducing control inputs, ultimately converging to an extremal solution of the AGHF \eqref{eqn:AGHF}, which minimizes the Action Functional.

\subsection{Trajectory optimization for pinned digit without obstacles}

\begin{figure}[h]
    \centering
    \begin{subfigure}{0.24\textwidth}
        \includegraphics[width=1\columnwidth]{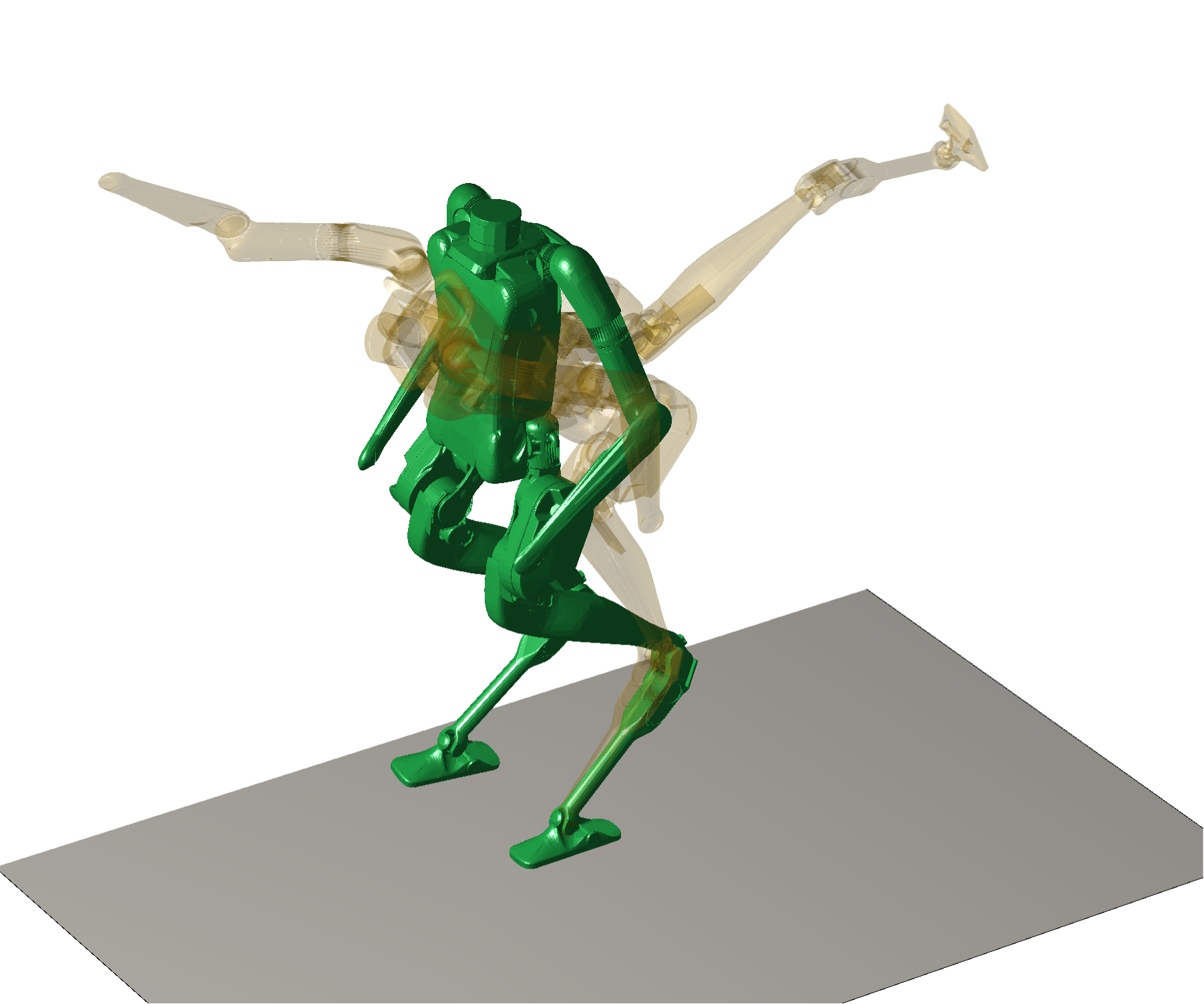}
    \end{subfigure}
    \hfill
    \begin{subfigure}{0.24\textwidth}
        \includegraphics[width=1\columnwidth]{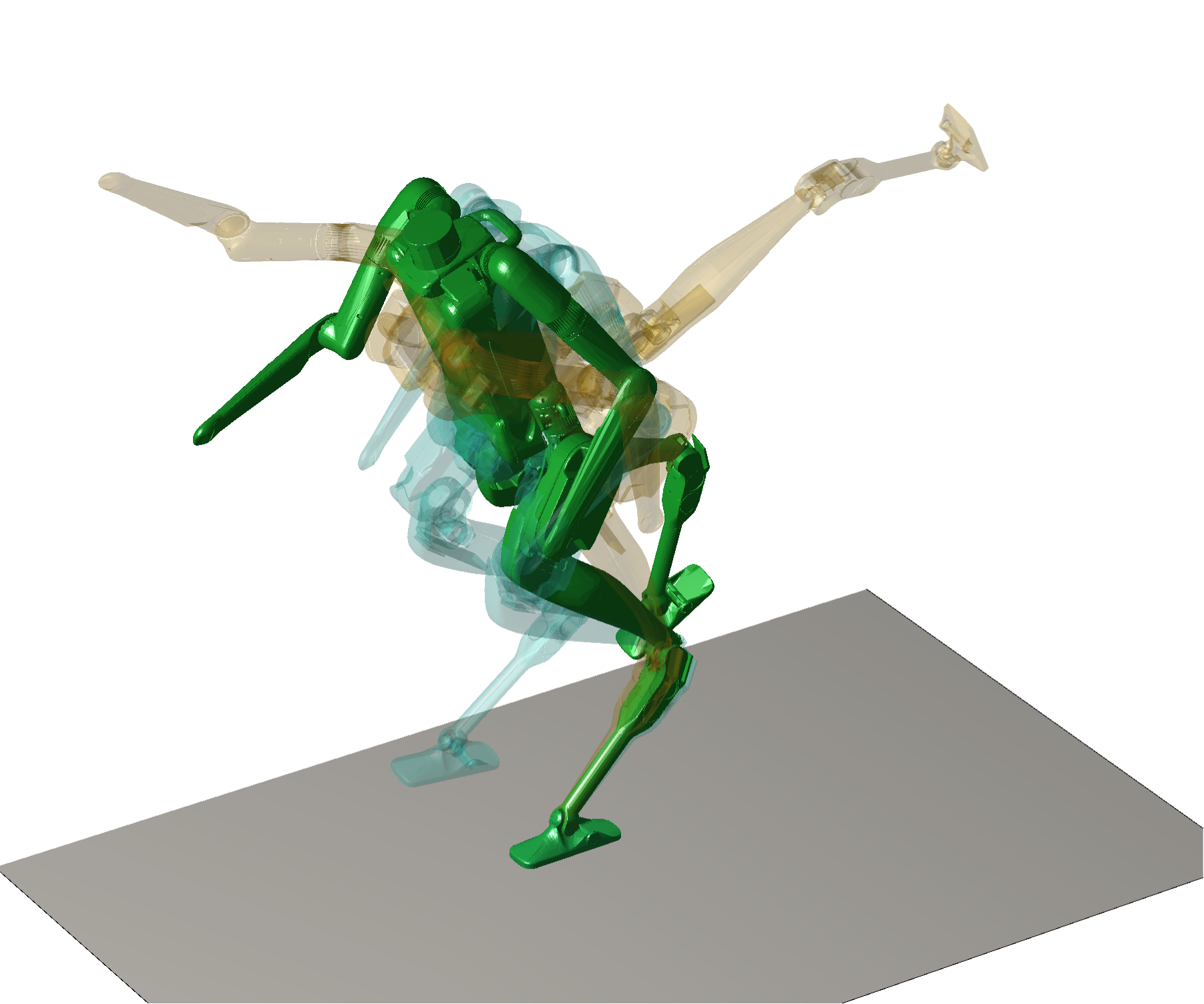}
    \end{subfigure}
    \hfill
    \begin{subfigure}{0.24\textwidth}
        \includegraphics[trim={0cm, 5cm, 0cm, 0cm},clip,width=1\columnwidth]{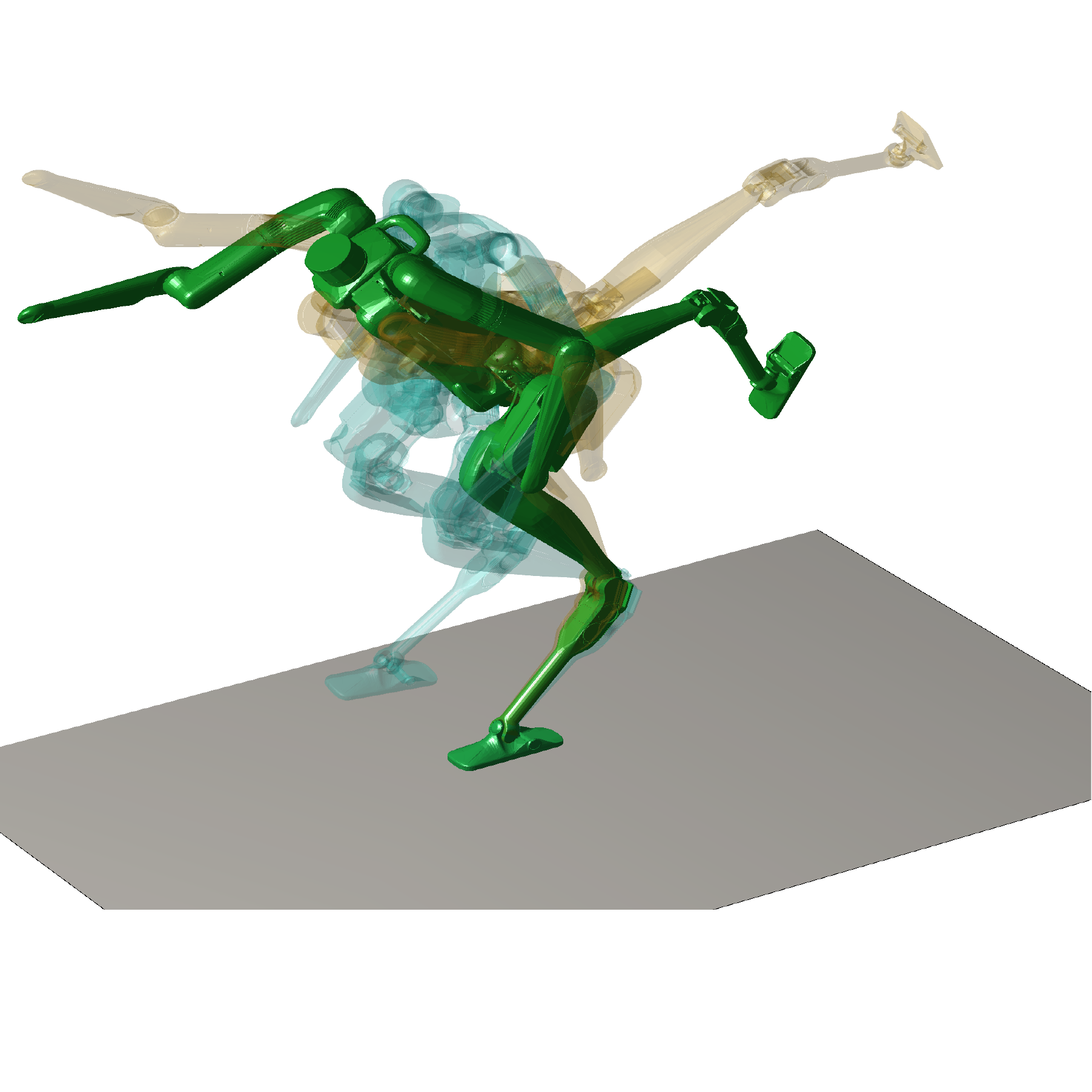}
    \end{subfigure}
    \hfill
    \begin{subfigure}{0.24\textwidth}
        \includegraphics[trim={0cm, 5cm, 0cm, 0cm},clip,width=1\columnwidth]{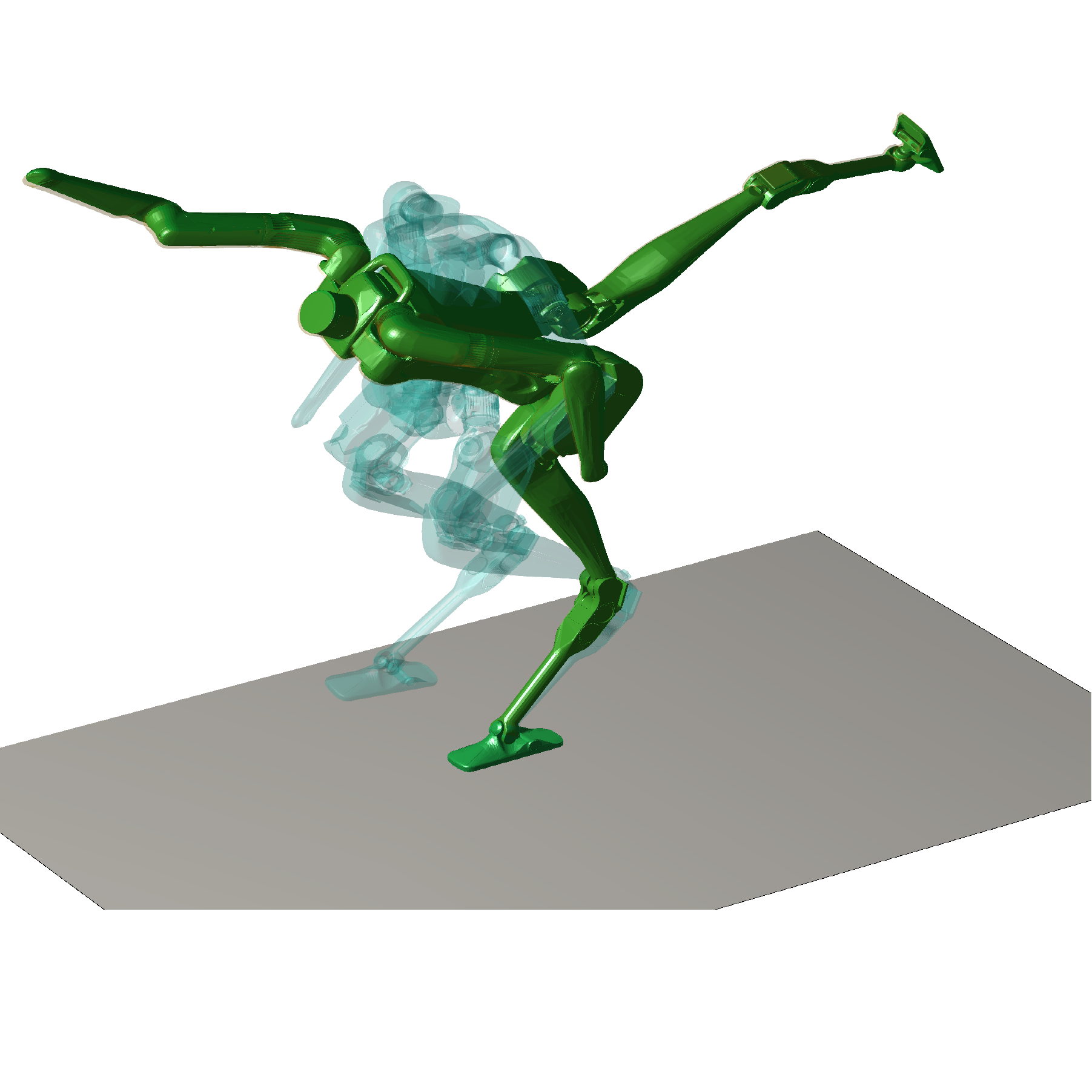}
    \end{subfigure}
    \caption{Snapshots of the forward simulated solution obtained by \methodname{} for Pinned Digit doing a yoga stretch. The intermediate poses in \textcolor{final_traj_green}{\textbf{green}}, the initial in \textcolor{init_pose_turq}{\textbf{turquoise}} and the final in \textcolor{final_pose_gold}{\textbf{gold}}.
    }
    \label{fig: digit multiple stills stretch}
\end{figure}

\begin{figure}[h]
    \centering
    \begin{subfigure}[b]{0.49\columnwidth}
        \centering
        \includegraphics[width=\columnwidth]{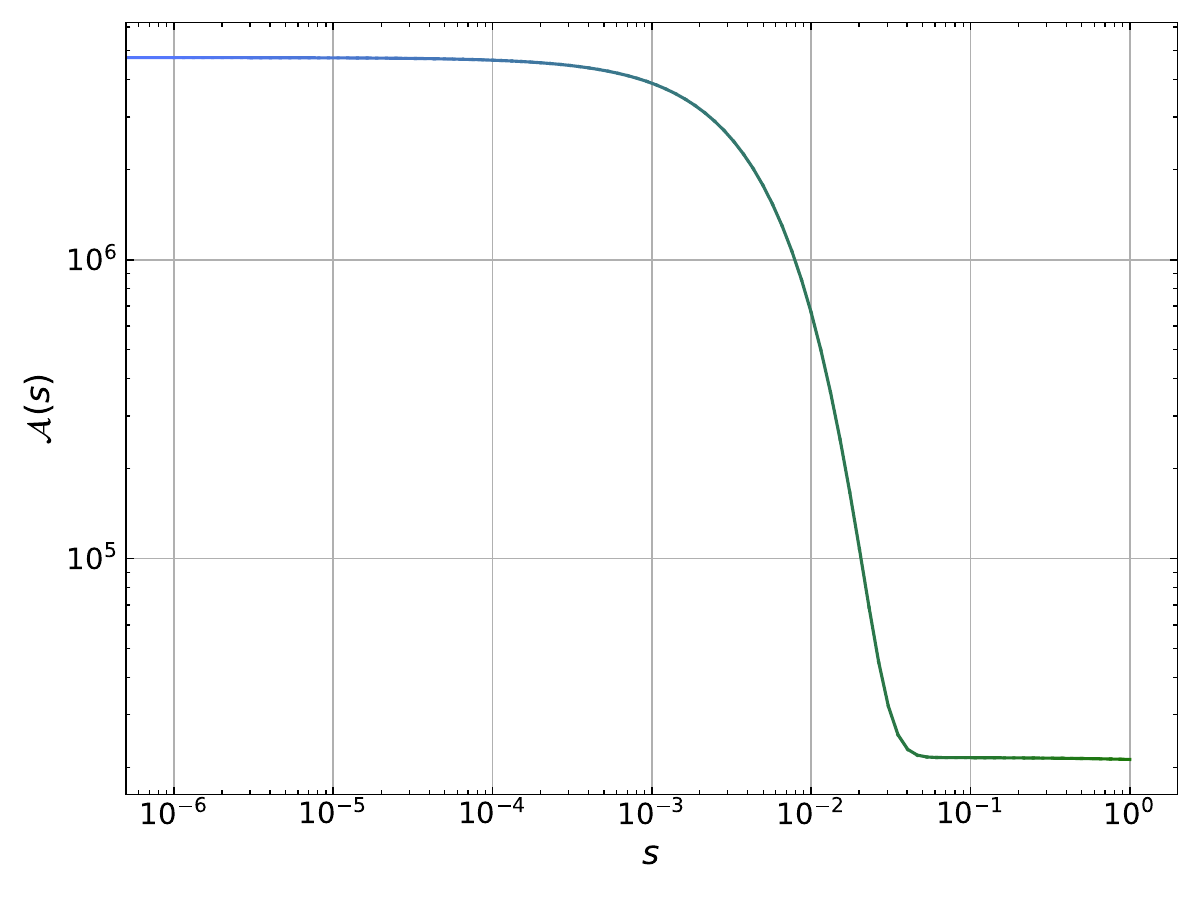}
        \caption{Convergence of the action functional ($\A(s)$) to a local minimum as the \methodname{} evolves along $s$ for the Digit yoga stretch scenario.}
    \end{subfigure}
    \hfill
    \begin{subfigure}[b]{0.49\columnwidth}
        \centering
        \includegraphics[width=\columnwidth]{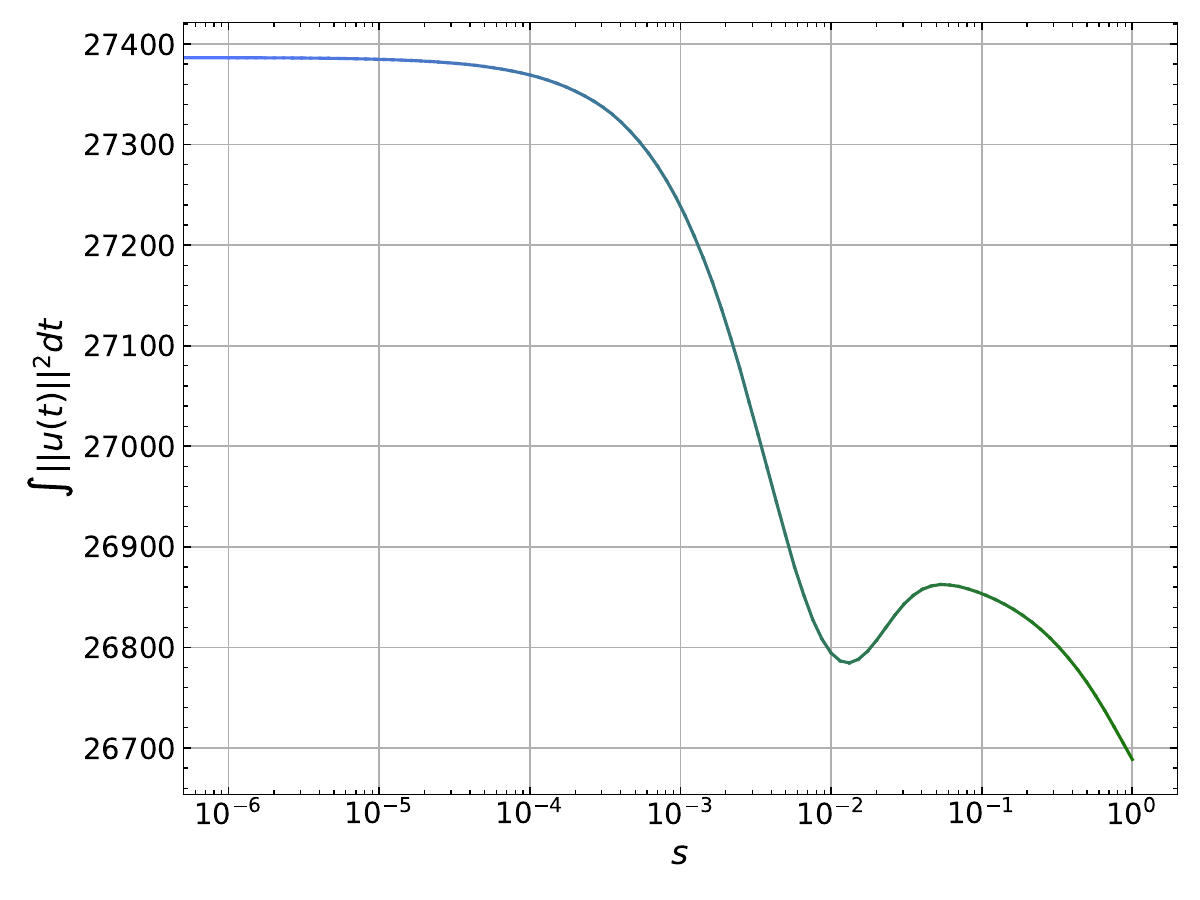}
        \caption{Evolution of $\int_0^T || u(t)||^2 dt $ as the \methodname{} evolves along $s$ for the Digit yoga stretch scenario.}
    \end{subfigure}
    \caption{Side-by-side figures showing the convergence of action functional and the evolution of $\int_0^T || u(t)||^2 dt $ for the Digit yoga stretch scenario.
    }
    \label{fig:combined stretch evolution action and u2}
\end{figure}

We consider two scenarios that correspond to taking a step and doing a yoga pose (Figure \ref{fig: digit multiple stills stretch}). 
In both experiments one of the feet is pinned to the ground. 
As before, we first run the grid search with the parameters in Tables \ref{table: Crocoddyl grid without obstacles} and \ref{table: aghf grid Digit without obstacles} for Crocoddyl and \methodname{}, respectively.
The best parameter set for \methodname{} was $s_{max}=1, k=10^7, p=6$.
The best parameter set for Crocoddyl was $w_x=0.001, w_u=0.0001, \delta t=0.001, w_{xf}=1000,$ and $\text{Initial Condition}=\text{``Rollout and Constant''}$.

Then, for each solver method, we find the single solver parameter set that yields the lowest time to solve when using a feedback controller with gains of $k_p=k_v=100$ in the step scenario.
After that, we ran all the scenarios sequentially ten times with the best parameter set and the mentioned controller gains.
Table \ref{table: comparison digit without obstacles} shows the results of these trajectory optimization problems. 
Both methods had a 100\% success rate, with \methodname{} being able to generate trajectories for the 22 DOF Digit biped in about 3s, two orders of magnitude faster than Crocoddyl.
However, Crocoddyl's trajectory had a smaller $\int_0^T || \Ufb(t)||^2 dt$. 
For the Pinned Digit Biped with the closed kinematic chain removed the an additional two sets of joints are made actuated to account for the removal of the closed loop kinematic chain. 
However, since these actuated joints are not present in the original Digit robot model, we cannot ascertain whether the computed trajectory adheres to the control limits of the actual hardware.

\begin{table}[h]
    \centering
        \begin{tabular}{ |c | c | c | c | c | c| }
        \hline
        \textbf{Method Name} & \textbf{Success Rate} & \boldsymbol{$\int_0^T || \Ufb(t)||^2 dt$} & \textbf{Solve Time [s]} \\
        \hline
        \textbf{\methodname{}} & 2/2 & 34570 ± 8418& 3.331 ± 1.437 \\
        \hline
        \textbf{Crocoddyl} & 2/2 & 5140 ± 1853& 225.2 ± 7.62\\
        \hline
        \end{tabular}
    \caption{A table showing the success rate, norm of the control effort ($\int_0^T || \Ufb(t)||^2 dt$) and solve time in seconds  for a set of Digit V3 trajectory optimization experiments with no obstacles}
    \label{table: comparison digit without obstacles}
\end{table}

\subsection{Trajectory Optimization for the Kinova Arm with obstacles}

\begin{figure}[h]
    \centering
    \begin{subfigure}[b]{0.49\columnwidth}
        \centering
        \includegraphics[trim={0cm, 0cm, 2cm, 0cm},clip,width=1\columnwidth]{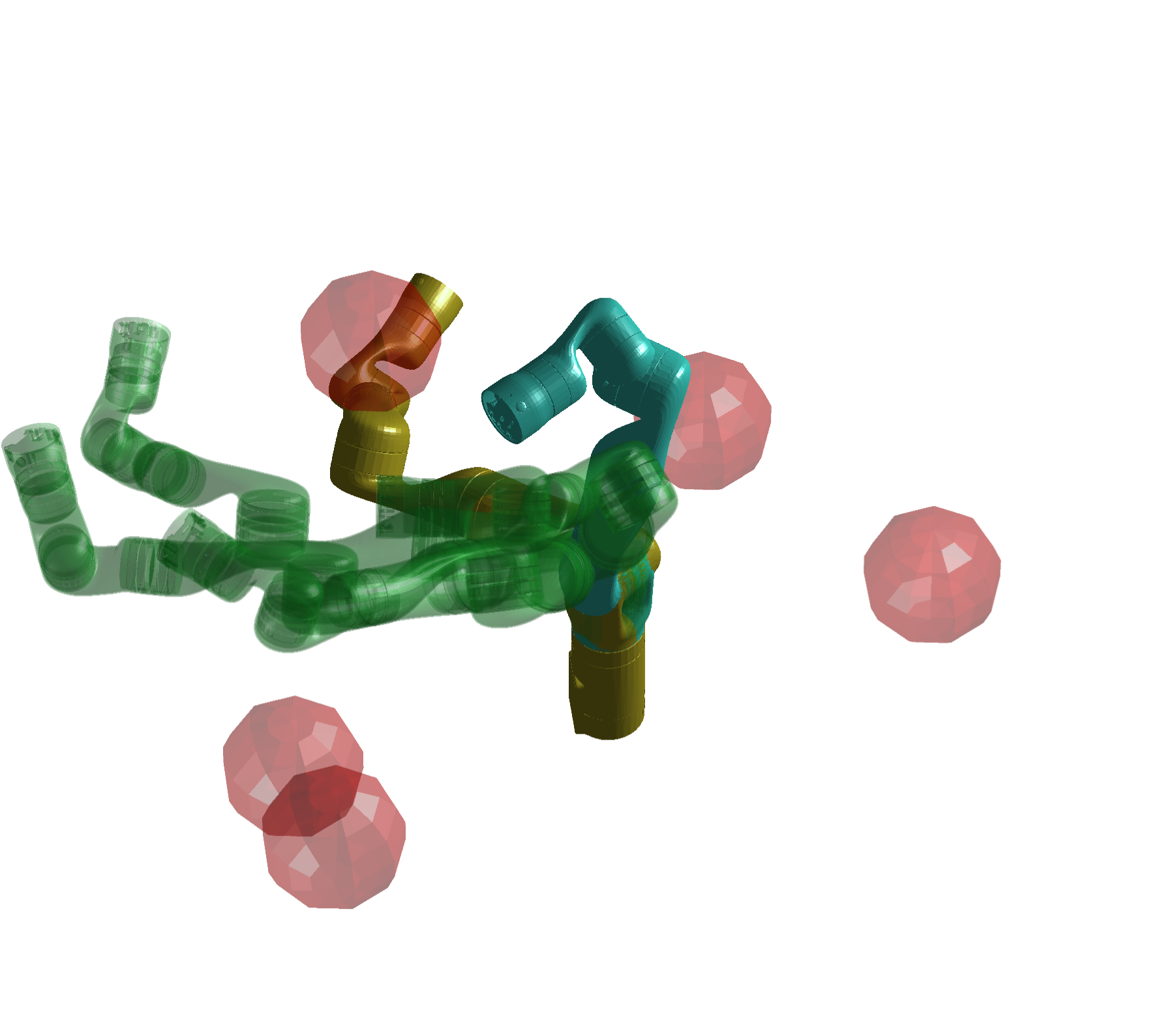}
    \end{subfigure}
    \hfill
    \begin{subfigure}[b]{0.49\columnwidth}
        \centering
        \includegraphics[trim={0cm, 0cm, 2cm, 0cm},clip,width=1\columnwidth]{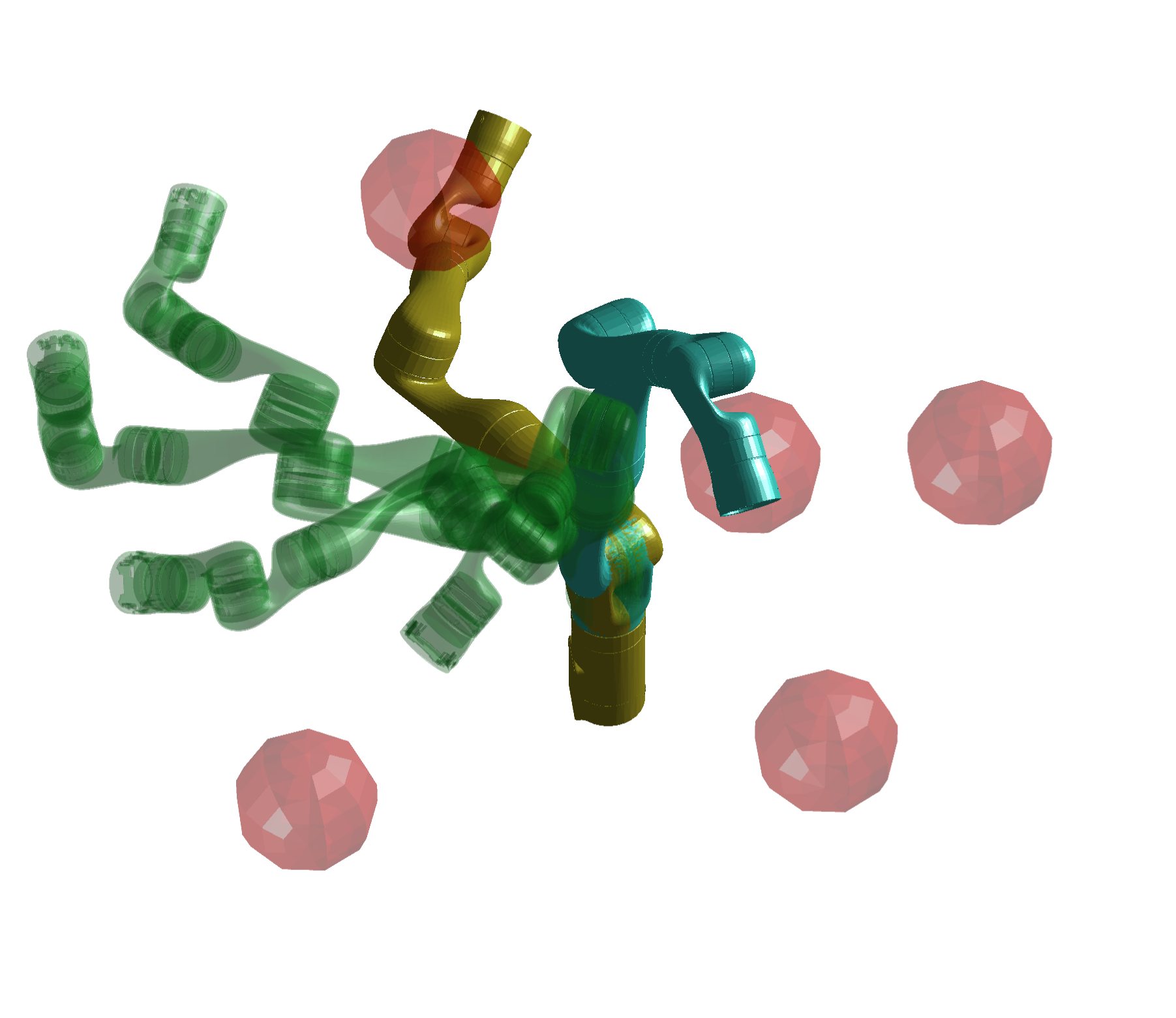}
        
    \end{subfigure}
    \caption{Two figures show trajectories generated by \methodname{} for the Kinova Gen3 to get from the initial poses in \textcolor{init_pose_turq}{\textbf{turquoise}} to the final poses in \textcolor{final_pose_gold}{\textbf{gold}} while avoiding the obstacles in \textcolor{obs_red}{{red}}. The intermediate poses in the paths it takes are shown in \textcolor{final_traj_green}{\textbf{green}}.
    }
    \label{fig:kinova exposure obstacles}
\end{figure}

In these experiments, we consider ten different scenarios where each one consists of a different $\x_0$, $\x_f$ similar to the one in Section \ref{subsec: kinova no obstacles}.
However, in each of these scenarios there are 5 obstacles in the arm's path that it must plan around to reach the goal.
We also start with an initial guess $\X_0$ that is not in collision with any of the obstacles.
As before, the average results for the best parameter sets per experiment are shown in Table \ref{table: comparison kinova with obstacles} where the controller gains were $k_p=k_v=10$.
The best parameter set for \methodname{} was $s_{max}=0.1, k=10^9, p=8, k_{cons}=10^9, c_{cons}=1$.
The best parameter set for Aligator was $w_x=0.001, w_u=0.0001, \delta t=0.001, w_{xf}=1,$ and $\text{Initial Condition}=\text{``Line and RNEA''}, \epsilon_{tol}=0.001, \mu_{init}=10^{-7}$.
In this setting \methodname{} is able to successfully avoid the obstacles and meet the goal in all the scenarios while generating trajectories much faster than Aligator.

Figure \ref{fig:kinova exposure obstacles} illustrates one of these scenarios.
Aligator had fewer successes than \methodname{}, but had a smaller control input.
Similar to the no obstacle case in Section \ref{subsec: kinova no obstacles}, the control inputs generated by \methodname{} still satisfied the control limits.

\begin{table}[H]
    \centering
    \begin{tabular}{ |c || c | c | c | c | c| }
    \hline
    \textbf{Method Name} & \textbf{Success Rate} & \boldsymbol{$\int_0^T || \Ufb(t)||^2 dt$} \textbf{of Successes} & \boldsymbol{$\int_0^T || \Ufb(t)||^2 dt$} \textbf{of Intersection} & \textbf{Solve Time[s]} & \textbf{Solve Time[s] of Intersection} \\
    \hline
    \textbf{\methodname{}} & 10/10 & 902.2 ± 122.8 & 903.8 ± 77.49 & 0.2968 ± 0.0492 & 0.3261 ± 0.03808 \\
    \hline
    \textbf{Aligator} & 6/10 & 194.3 ± 78.95 & 194.3 ± 78.95 & 75.84 ± 9.405 & 75.84 ± 9.405 \\
    \hline
    \end{tabular}
    \caption{A table showing the success rate, norm of the control effort ($\int_0^T || \Ufb(t)||^2 dt$) and Solve Time in seconds  for Kinova Gen3 trajectory optimization experiment in the presence of obstacles.
    The 'Intersection' columns report metrics for trials where both \methodname{} and Aligator succeeded, while the 'Successes' columns provide metrics based solely on the successful trials of each method.
    }
    \label{table: comparison kinova with obstacles}
\end{table}

\subsection{Trajectory optimization for pinned digit with obstacles}
In this section we consider four experiments that consist of the pinned Digit taking a single step over a single half-sphere obstacle of four different radii, with the smallest and largest spheres having radii of $1$cm and $20.3$cm respectively.
Once again we start with an initial guess $\X_0$ that is not in collision with any of the obstacles.
The average results for the best parameter sets are shown in Table \ref{table: comparison digit with obstacles} where the controller gains were $k_p=k_v=100$.
The best parameter set for \methodname{} was $s_{max}=1.0, k=10^7, p=6, k_{cons}=10^5, c_{cons}=200$.
For Aligator none of the parameter sets yielded a success when the allowable final state error $\epsilon = 0.05$ as with all the other experiments.

Here we see that \methodname{} is able to successfully generate trajectories to get Digit to step over the obstacles in less than 6s on average.
Whereas Aligator is unable to succeed in any of the trials due to it's final state being too far from the goal state.

\begin{table}[H]
    \centering
    \begin{tabular}{ |c | c | c | c | }
    \hline
    \textbf{Method name} & \textbf{Success Rate} & \boldsymbol{$\int_0^T || \Ufb(t)||^2 dt$} & \textbf{Solve time [s]} \\
    \hline
    \textbf{\methodname{}} & 4/4 & 43633 ± 904.6 & 5.285 ± 0.2262 \\
    \hline
    \textbf{Aligator} & 0/4 & DNF & DNF \\
    \hline
    \end{tabular}
    \caption{A table showing the success rate, norm of the control effort ($\int_0^T || \Ufb(t)||^2 dt$) and Solve Time in seconds  for Digit trajectory optimization experiment where Digit steps over different sized obstacles when the threshold for success $\epsilon$ is set to 0.05. We denote the results where one of the methods was unable to generate a successful solution as DNF (Did Not Find).}
    \label{table: comparison digit with obstacles}
\end{table}

Because Aligator had no successes with the regular error threshold for success ($\epsilon$ = 0.05), we reran the same experiment of taking a step, but with a larger error threshold ($\epsilon$ = 0.25). 
The results for this experiment are shown in Table \ref{table: comparison digit with obstacles th 0.25}.
The results show that \methodname{} still outperforms Aligator in terms of time to solve the optimal control problem.
The best parameter set for Aligator was $w_x=1.0, w_u=0.01, \delta t=0.01, w_{xf}=10^{-6},$ and $\text{Initial Condition}=\text{``Zeros''}, \epsilon_{tol}=0.001, \mu_{init}=10^{-8}$.

\begin{table}[H]
    \centering
    \begin{tabular}{ |c | c | c | c | c | c| }
    \hline
    \textbf{Method name} & \textbf{Success Rate} & \boldsymbol{$\int_0^T || \Ufb(t)||^2 dt$} & \boldsymbol{$\int_0^T || \Ufb(t)||^2 dt$} \textbf{of Intersection} & \textbf{Solve Time [s]} & \textbf{Solve Time of Intersection [s]} \\
    \hline
    \textbf{\methodname{}} & 4/4 & 43633 ± 904.6 & 42964 ± 0 & 5.285 ± 0.2262 & 5.335 ± 0.2984 \\
    \hline
    \textbf{Aligator} & 1/4 & 68306 ± 0 & 68306 ± 0 & 187.6 ± 1.571 & 187.6 ± 1.571 \\ 
    \hline
    \end{tabular}
    \caption{
    A table showing the success rate, norm of the control effort ($\int_0^T || \Ufb(t)||^2 dt$) and Solve Time in seconds  for Digit trajectory optimization experiment where Digit steps over different sized obstacles when the threshold for success $\epsilon$ is set to 0.25.
    }
    \label{table: comparison digit with obstacles th 0.25}
\end{table}

\section{Conclusion and Limitations}
\label{sec:conclusion}

This work proposes \methodname{}, a method for rapid trajectory optimization leveraged by solving the AGHF PDE.
The AGHF PDE poses the trajectory generation problem as the solution to a PDE that evolves an initial trajectory that may not be dynamically feasible into some final trajectory that is dynamically feasible while minimizing the magnitude of the control inputs of the final trajectory.
\methodname{} applies pseudospectral collocation in conjunction with spatial vector algebra to rapidly solve the AGHF PDE enabling it to scale to high-dimensional robot systems.
This paper demonstrates that \methodname{} can generate trajectories for high-dimensional systems much faster than state-of-the-art trajectory optimization methods, with \methodname{} generating trajectories for the Digit biped on the order of 4 seconds.

 \methodname{} has its limitations: because the action functional \eqref{eqn:action functional k} incorporates dynamic constraints through a penalty term, the AGHF solution does not minimize the control inputs to the fullest extent possible.
Additionally, we have demonstrated \methodname{} on legged systems by assuming that ground contact can always be maintained (i.e., we assumed there was infinite friction). 
Future work will extend this formulation to systems while checking to see if contact can be maintained.
\section{Acknowledgements}
\label{sec: acknowledgements}

This work was supported by AFOSR MURI
FA9550-23-1-0400, and the Automotive Research Center (ARC).
The authors thank Dr. Y. Fan, Professor S. Liu and Professor A. Bellabas for their insights on the AGHF.
The authors also thank Dr. S. Singh and Professor P. Wensing for their insights on the second derivatives of dynamics.
The authors also want to thank Dr. J. Liu for his support at the early stages of this work.

\renewcommand{\bibfont}{\normalfont\small}
{\renewcommand{\markboth}[2]{}
\printbibliography}

\pagebreak

\appendices
\section{Proof of Theorem \ref{thm: Analytical PDE expression}}
\label{sec: AGHF RHS Appendix}
\begin{proof}
     The Affine Geometric Heat Flow (AGHF) Equation is given by 
     \begin{align}
    \label{eqn:AGHF_appendix}
       \dxds(t,s)  = \Ginv(\X(t,s)) \bigg(\ddt \dldxds(\X_s(t),\dot{\X}_s(t)) - \dldxs(\X_s(t),\dot{\X}_s(t))\bigg)
    \end{align}
    Let
     \begin{align}
    \label{eqn:lagrangian_appendix}
    L(\X_s(t),\dot{\X}_s(t)) = \left(\dot{\X}_s(t) - F_d(\X_s(t))\right)^T G(\X_s(t)) \left(\dot{\X}_s(t)- F_d(\X_s(t))\right)
    \end{align}
    where  
    \begin{align}
    \label{eqn:G(x)_appendix_out}
    \begin{split}
        & G(\X_s(t)) = (\bar{F}(\X_s(t))^{-1})^T K \bar{F}(\X_s(t))^{-1} \\
        & G(\X_s(t)) = \underbrace{\begin{bmatrix}
        I_{N \times (2N-m)} & 0_{N \times m}\\
        0_{N \times (2N-m)} & B^{-1}H(\xpone(t,s)) \end{bmatrix}^T }_{(\bar{F}(\X_s(t))^{-1})^T}
        \underbrace{\begin{bmatrix}
        kI_{N \times N} & 0_{N \times N} \\
        0_{N \times N} & I_{N \times N}
        \end{bmatrix}}_{K}
        \underbrace{\begin{bmatrix}
        I_{N \times (2N-m)} & 0_{N \times m}\\
        0_{N \times (2N-m)} & B^{-1}H(\xpone(t,s))
        \end{bmatrix} }_{(\bar{F}(\X_s(t))^{-1})}
    \end{split}
    \end{align}
    We assume, without loss of generality (WLOG), that $B = I$. 
    Under this assumption, multiplying the matrices in \eqref{eqn:G(x)_appendix_out} yields
    \begin{align}
    G(\X_s(t)) = \begin{bmatrix}
        kI_{N \times N} & 0_{N \times N} \\
        0_{N \times N} & \HT H
    \end{bmatrix}   
    \end{align}

    From here for conciseness we drop the dependence on $(t,s)$ for the following equations (i.e. $\X(t,s)$ is  denoted by $\X$).
    We also drop the dependence on the $\X$ terms for the dynamics functions~(i.e. $H(\xpone)$ is denoted by just $H$ and so on).
    
    To compute \eqref{eqn:AGHF_appendix}, we need to compute the derivatives $\ddt \dldxd$  and $\dldx$. 
    We begin by showing that $\ddt \dldxd = \Omega_1$.
    First we must compute the derivative $\dldxd$,
    \begin{align}
        \dldxd = 2 G (\Xd - \Fd) =
        2 \underbrace{
        \begin{bmatrix}
            kI_{N \times N} & 0_{N \times N} \\
            0_{N \times N} & \HT H
        \end{bmatrix}
        }_{G}
        \bigg(
        \underbrace{
        \begin{bmatrix}
        \xdpone \\
        \xdptwo
        \end{bmatrix}
        }_{\Xd}
        -
        \underbrace{
        \begin{bmatrix}
        \xptwo \\
        - \Hinv C
        \end{bmatrix}
        }_{\Fd}
        \bigg)
        = 2
        \begin{bmatrix}
        k(\xdpone - \xptwo) \\
        \HT H \xdptwo + \HT C
        \end{bmatrix}
    \end{align}
    taking the time derivative of this yields:
    
    \begin{align}
        \ddt \dldxd = 
        2 \begin{bmatrix}  k(\Xdd_{P1} - \Xd_{P2}) \\ 
            (\HdotT H + \HT \Hdot) \Xd_{P2} + \HT H \Xdd_{P2} + \HdotT C + \HT \Cdot
        \end{bmatrix}
    \end{align}

    and multiplying by $\Ginv$ yields

    \begin{align}
        \Omega_1 = \Ginv \bigg(\ddt \dldxd \bigg) = 
        2 \begin{bmatrix}  \Xdd_{P1} - \Xd_{P2} \\ 
            \HTHinv \big((\HdotT H + \HT \Hdot) \Xd_{P2} + \HT H \Xdd_{P2} + \HdotT C + \HT \Cdot \big)
        \end{bmatrix}
    \end{align}

    Next we show that $\Ginv \dldx = \Omega_2 - \Omega_3 + \Omega_4$ 

    First, taking the derivative of \eqref{eqn:lagrangian_appendix} wrt. $\X$, and taking transposes to ensure that the resultant derivatives are column vectors we obtain:
    \begin{align}
         \dldx = -\dFdTdx G (\Xd - \Fd) + (\Xd-\Fd)^T \dGdx (\Xd - \Fd) + \bigg( (\Xd-\Fd)^T G (-\dFddx) \bigg)^T
    \end{align}

    expanding the first and third terms yields

    \begin{align}
        \dldx = - \dFdTdx G \Xd + \dFdTdx G \Fd + (\Xd-\Fd)^T \dGdx (\Xd - \Fd) + \bigg(-\Xd^T G \dFddx \bigg)^T + \bigg(\Fd^T G \dFddx \bigg)^T
    \end{align}

    Since G is symmetric (because $kI$ and $\HT H$ are symmetric) the last two terms can be simplified yielding:

    \begin{align}
        \dldx = - \dFdTdx G \Xd + \dFdTdx G \Fd + (\Xd-\Fd)^T \dGdx (\Xd - \Fd) -\dFdTdx G \Xd + \dFdTdx  G \Fd
    \end{align}

    Grouping terms and simplifying yields

    \begin{align}
        \label{eqn: dldx_appendix}
        \dldx = 2\dFdTdx G \Fd - 2\dFdTdx G \Xd + (\Xd-\Fd)^T \dGdx (\Xd - \Fd)
    \end{align}

    Let $\FDuzero = -\Hinv C$.
    Next we derive the constituent terms in \eqref{eqn: dldx_appendix}.
    The first term is given by:
    
    \begin{align}
        2 \dFdTdx G \Fd = 
        \renewcommand{\arraystretch}{1.5}
        2
        \underbrace{
        \begin{bmatrix}
            0_{N \times N} & \dFDTdxpone \\
            I_{N \times N} & \dFDTdxptwo
        \end{bmatrix}
        }_{\dFdTdx}
        \underbrace{
        \begin{bmatrix}
            kI_{N \times N} & 0_{N \times N} \\
            0_{N \times N} & \HT H
        \end{bmatrix}
        }_{G}
        \underbrace{
        \begin{bmatrix}
        \xptwo \\
        - \Hinv C
        \end{bmatrix}
        }_{\Fd}
        =
        2\begin{bmatrix}
            - \dFDTdxpone \HT C \\
            k\xptwo - \dFDTdxptwo \HT C
        \end{bmatrix}
    \end{align}

    and the second term:
    \begin{align}
    \renewcommand{\arraystretch}{1.5}
        2\dFdTdx G \Xd = 2
        \underbrace{
        \begin{bmatrix}
            0_{N \times N} & \dFDTdxpone \\
            I_{N \times N} & \dFDTdxptwo
        \end{bmatrix}
        }_{\dFdTdx}
        \underbrace{
        \begin{bmatrix}
            kI_{N \times N} & 0_{N \times N} \\
            0_{N \times N} & \HT H
        \end{bmatrix}
        }_{G}
        \underbrace{
        \begin{bmatrix}
        \xdpone \\
        \xdptwo
        \end{bmatrix}
        }_{\Xd}
        = 2
        \begin{bmatrix}
            \dFDTdxpone \HT H \xdptwo \\
            k\xdpone + \dFDTdxptwo \HT H \xdptwo
        \end{bmatrix}
    \end{align}

    Collecting and rearranging the first 2 terms of \eqref{eqn: dldx_appendix} yields:
    \begin{align}
    \renewcommand{\arraystretch}{2.0}
        2\dFdTdx G \Fd - 2\dFdTdx G \Xd = 
        \begin{bmatrix} 
            -\dFDTdxpone \HT \bigg(2C + 2H \dot{\X}_{P2} \bigg) \\ 
            -\dFDTdxptwo \HT \bigg(2C + 2H \dot{\X}_{P2} \bigg) 
        \end{bmatrix} 
        -
        \renewcommand{\arraystretch}{1.0}
        \begin{bmatrix}
            \vspace{3.1pt}
            \textbf{0} \\
             2k(\Xd_{P1} -\X_{P2})\\
        \end{bmatrix}
    \end{align}

    and multiplying them by $\Ginv$ gives us $\Omega_2$ and $\Omega_3$:
    \renewcommand{\arraystretch}{2}
    \begin{align}
         \Ginv \bigg(\dFdTdx G \Fd +  \bigg(\Fd^T G \dFddx \bigg)^T - 2\dFdTdx G \Xd \bigg) = 
         \underbrace{
         \begin{bmatrix} -\frac{1}{k}I_{N \times N} \dFDTdxpone \HT \bigg(2C + 2H \dot{\X}_{P2} \bigg) \\ 
            -(\HT H)^{-1} \dFDTdxptwo \HT \bigg(2C + 2H \dot{\X}_{P2} \bigg) \bigg) 
        \end{bmatrix}
        }_{\Omega_2}
        - 
        \renewcommand{\arraystretch}{1.0}
        \underbrace{
        \begin{bmatrix}
            \vspace{3.1pt}
            \textbf{0} \\
            \vspace{3.1pt}
             2k (\HT H)^{-1} (\dot{\X}_{P1} -{\X}_{P2})\\
        \end{bmatrix}
        }_{\Omega_3}
    \end{align}

    The third term of \eqref{eqn: dldx_appendix} is given by:

    \begin{align}
    \label{eqn: dGdx_term_appendix}
    \renewcommand{\arraystretch}{1.0}
        (\Xd-\Fd)^T \dGdx (\Xd - \Fd) = 
            \vspace{1pt}
            \left(\Xd - \Fd\right) ^T
             \begin{bmatrix} 0_{N \times N \times 2N} & \\
                             & \dHTHdx \end{bmatrix}
             \vspace{1pt}
             \left( \Xd - \Fd \right),
    \end{align}

    where $\dHTHdxi$ be the partial derivative of $\HT H$ wrt. the $i$th element of the $2N \times 1$ column vector $\X$.
    Doing the vector-tensor-vector multiplication in \eqref{eqn: dGdx_term_appendix} yields an $2N \times 1$ column vector of the following form:

    \begin{align}
    \renewcommand{\arraystretch}{1.0}
    \label{eqn: dGdx_quadratic_form}
        (\Xd-\Fd)^T \dGdx (\Xd - \Fd) = 
         \begin{bmatrix}
            \left(\Xd - \Fd\right) ^T
             \begin{bmatrix} 0_{N \times N} & \\
                             & \dHTHdxone \end{bmatrix}
             \left( \Xd - \Fd \right) \\
            \vdots \\
            \left(\Xd - \Fd\right) ^T
             \begin{bmatrix} 0_{N \times N} & \\
                             & \dHTHdxtwoN \end{bmatrix}
             \left( \Xd - \Fd \right)
     \end{bmatrix}
    \end{align}

    and applying \eqref{eqn: Fd} to \eqref{eqn: dGdx_quadratic_form} yields:

    \begin{align}
    \renewcommand{\arraystretch}{1.0}
        (\Xd-\Fd)^T \dGdx (\Xd - \Fd) = 
         \begin{bmatrix}
            \left(\xdptwo - \FDuzero \right) ^T
             \dHTHdxone
             \left(\xdptwo - \FDuzero \right) \\
            \vdots \\
            \left(\xdptwo - \FDuzero \right) ^T
             \dHTHdxtwoN
             \left( \xdptwo - \FDuzero \right)
     \end{bmatrix}
    \end{align}

    Notice that since $H$ is only a function of $\xpone$, then
    \begin{align}
    \renewcommand{\arraystretch}{1.0}
         \begin{bmatrix}
             \dHTHdxNone \\
            \vdots \\
             \dHTHdxtwoN
     \end{bmatrix}
     =
     \begin{bmatrix}
             0 \\
            \vdots \\
             0
     \end{bmatrix}
    \end{align}

    and 

    \begin{align}
    \renewcommand{\arraystretch}{1.0}
    \label{eqn: dGdx_quadratic_form_zeros}    
        (\Xd-\Fd)^T \dGdx (\Xd - \Fd) = 
        \begin{bmatrix}
        \renewcommand{\arraystretch}{2.0}
         \begin{bmatrix}
            \left(\xdptwo - \FDuzero \right) ^T
             \dHTHdxone
             \left(\xdptwo - \FDuzero \right) \\
            \vdots \\
            \left(\xdptwo - \FDuzero \right) ^T
             \dHTHdxN
             \left( \xdptwo - \FDuzero \right) \\
     \end{bmatrix}\\
     \renewcommand{\arraystretch}{1.0}
     \begin{bmatrix}
             0 \\
            \vdots \\
             0
     \end{bmatrix}
    \end{bmatrix}
    =
    \renewcommand{\arraystretch}{2.0}
    \begin{bmatrix}
            \left(\xdptwo - \FDuzero \right) ^T
             \dHTHdxpone
             \left(\xdptwo - \FDuzero \right) \\
             \textbf{0}
    \end{bmatrix}
    \end{align}

    Next we show how to further simplify \eqref{eqn: dGdx_quadratic_form_zeros} to yield $\Omega_4$.
    First, let $\beta_i$ be defined as the $i$th entry of $(\Xd-\Fd)^T \dGdx (\Xd - \Fd)$ given by:

    \begin{align}
    \renewcommand{\arraystretch}{1.0}
    \label{eqn: beta_i}
        \beta_i = 
            \left(\xdptwo - \FDuzero \right) ^T
             \dHTHdxi
             \left( \xdptwo - \FDuzero \right)
    \end{align}

    evaluating $\dHTHdxi$ and expanding \eqref{eqn: beta_i} gives:
    \begin{align}
    \renewcommand{\arraystretch}{1.0}
        \beta_i = 
            \left(\xdptwo - \FDuzero \right) ^T
             \dHTdxi H
             \left( \xdptwo - \FDuzero \right) 
             +
             \left(\xdptwo - \FDuzero \right) ^T
             \HT \dHdxi
             \left( \xdptwo - \FDuzero \right)
    \end{align}

    Notice $\beta_i$ is a scalar, and so are each of its constituent terms. 
    Therefore the second term of $\beta_i$ can be expressed in the following way

    \begin{align}
    \label{eqn: beta_i_second_term}
    \begin{split}        
             & \left(\xdptwo - \FDuzero \right) ^T
             \HT \dHdxi
             \left( \xdptwo - \FDuzero \right) = 
             \bigg(\left(\xdptwo - \FDuzero \right) ^T
             \HT \dHdxi
             \left( \xdptwo - \FDuzero \right) \bigg) ^T \\
             & \hspace{5.55cm} =
             \left(\xdptwo - \FDuzero \right) ^T
             \dHTdxi H
             \left( \xdptwo - \FDuzero \right)
     \end{split}
    \end{align}

    Using \eqref{eqn: beta_i_second_term} to simplify $\beta_i$ gives
    \begin{align}
        \beta_i = 
            2 \left(\xdptwo - \FDuzero \right) ^T
             \dHTdxi H
             \left( \xdptwo - \FDuzero \right)
             =
             2 \begin{bmatrix} \dHdxi \left(\xdptwo - \FDuzero \right) \end{bmatrix}^T H
             \left( \xdptwo - \FDuzero \right)
    \end{align}

    Applying this to \eqref{eqn: dGdx_quadratic_form_zeros} yields
    \begin{align}
    \label{eqn: dGdx_quadratic_form_final}
        (\Xd-\Fd)^T \dGdx (\Xd - \Fd) = 
        \begin{bmatrix}
            2 \begin{bmatrix} \dHdxpone \left(\xdptwo - \FDuzero \right) \end{bmatrix}^T
             H
             \left(\xdptwo - \FDuzero \right) \\
             \textbf{0} \\
        \end{bmatrix}
    \end{align}
    
    Note that to ensure that \eqref{eqn: dGdx_quadratic_form_final} evaluates to the same terms as \eqref{eqn: dGdx_quadratic_form_zeros} the 3rd dimension of the tensor $\dHdxpone$ must be used in the tensor-vector multiplication with $\xdptwo - \FDuzero$.
    
    Lastly, multiplying \eqref{eqn: dGdx_quadratic_form_final} by $\Ginv$ gives us $\Omega_4$:

    \begin{align}
    \renewcommand{\arraystretch}{2.0}
        \Ginv (\Xd-\Fd)^T \dGdx (\Xd - \Fd) = 
         \begin{bmatrix}
            2 \frac{1}{k}
            \begin{bmatrix} \dHdxpone \left(\xdptwo - \FDuzero \right) \end{bmatrix}^T
             H
             \left(\xdptwo - \FDuzero \right) \\
             \textbf{0} \\
        \end{bmatrix}
        = \Omega_4
    \end{align} 

    Combining these terms we have

    \begin{align}
       \dxds  = \Ginv \bigg(\ddt \dldxds - \dldxs \bigg) = \Omega_1 - (\Omega_2 - \Omega_3 + \Omega_4) = \Omega \bigg( \X, \Xd, \Xdd, k \bigg)
    \end{align}
    
\end{proof}
\section{Proof of Theorem \ref{thm: PDE jacobian} }
\label{sec: AGHF Jacobian Appendix}

\begin{proof}
    Let $\Jpi(s)$ be the Jacobian of $\frac{d \psX_i}{ds}(s)$ with respect to $\psX_i (s)$ given by:
    \begin{equation}
    \label{eqn: jacobian appendix}
    \begin{split}
        & \Jpi = \frac{d \big(\frac{d \psX_i}{ds}(s) \big)}{d\psX_i (s)} = \frac{ d\Omega \big( \psX_i(s),  [D\psX]_i(s), [D^2\psX]_i(s), k \big)}{d\psX_i (s)}.
    \end{split}
    \end{equation}
    Applying the chain rule for multivariable functions we obtain
    \begin{equation}
    \begin{split}
        & \Jpi = \frac{d \Omega}{d \psX_i} \frac{d \psX_i}{d \psX_i} + \frac{d \Omega}{d [D\psX]_i} \frac{d [D\psX]_i}{d \psX_i} + \frac{d \Omega}{d [D^2\psX]_i} \frac{d [D^2\psX]_i}{d \psX_i}
    \end{split}
    \end{equation}

    From Theorem \ref{thm: Analytical PDE expression} we have that
    \begin{equation}
    \label{eqn:aghf omega_appendix}
    \begin{split}
        & \frac{\partial \X}{\partial s} = \Omega \bigg( \X, \dot{\X}, \ddot{\X}, k \bigg) = \Omega_1 - (\Omega_2 - \Omega_3 + \Omega_4),
    \end{split}
    \end{equation}

    And recall that $\psX_i(s) = \X^{T}(t_i,s)$.
    Using these two results, we have that
    \begin{align}
    \label{eqn: d_omega_d_psi_appendix}
    \begin{split}
         \frac{d \Omega}{d \psX_i} = \frac{d \Omega}{d \X(t_i,s)}
         = \frac{d \Omega_1}{d\X} - (\dOmegatwodx - \dOmegathreedx +\dOmegafourdx)
    \end{split}
    \end{align}

    Computing the derivatives of each of the $\Omega$ terms yields the following:
    \begin{align}
    \begin{split}
        \frac{d \Omega_1}{d \X} 
        = 2 \begin{bmatrix} 0_{N \times N} & 0_{N \times N} \\ 
         \dHTHinvdxpone \gamma + \HTHinv \dgammadxpone & \HTHinv \dgammadxptwo  
        \end{bmatrix}
    \end{split}
    \end{align}

    \begin{align}
    \begin{split}
        \frac{d \Omega_2}{d \X}
        = \begin{bmatrix} \frac{1}{k}I_{N \times N} \dalphaonedxpone & \frac{1}{k}I_{N \times N} \dalphaonedxptwo \\ 
         \dHTHinvdxpone \alpha_2 + \HTHinv \dalphatwodxpone & \HTHinv \dalphatwodxptwo  
        \end{bmatrix}
    \end{split}
    \end{align}

    \begin{align}
    \begin{split}
        \frac{d \Omega_3}{d \X}
        = 2k \begin{bmatrix} 0_{N \times N} & 0_{N \times N} \\ 
         \dHTHinvdxpone (\xdpone - \xptwo) & -\HTHinv I_{N \times N}  
        \end{bmatrix}
    \end{split}
    \end{align}

    \begin{align}
    \begin{split}
        \frac{d \Omega_4}{d \X}
        = \begin{bmatrix}  \frac{2}{k}I_{N \times N}\dGammadxpone & \frac{2}{k}I_{N \times N} \dGammadxptwo \\ 
         0_{N \times N} & 0_{N \times N}  
        \end{bmatrix}
    \end{split}
    \end{align}

    where,
    \begin{align}
    \begin{split}
    \label{gamma_appendix}
        \gamma = (\HdotT H + \HT \Hdot) \dot{\X}_{P2} + \HT H \ddot{\X}_{P2} + \HdotT C + \HT \Cdot
    \end{split}
    \end{align}

    \begin{align}
    \begin{split}
        \alpha_1 = -\dFDTdxpone \HT \bigg(2C + 2H \dot{\X}_{P2} \bigg)
    \end{split}
    \end{align}

    \begin{align}
    \begin{split}
        \alpha_2 = -\dFDTdxptwo  \HT \bigg(2C + 2H \dot{\X}_{P2} \bigg)
    \end{split}
    \end{align}

    \begin{align}
    \begin{split}
        \Gamma = (\xdptwo - \FDuzero)^T \dHTdxpone H (\xdptwo - \FDuzero)
    \end{split}
    \end{align} 

    \begin{align}
    \begin{split}
        & \dgammadxpone = \dHdotTdxpone \bigg(H \xdptwo + C \bigg) + \Hdot^T \bigg( \dHdxpone \xdptwo + \dCdxpone \bigg) + \dHTdxpone \bigg( \Hdot \xdptwo + H \xddptwo + \Cdot \bigg) \\
        & + \HT \bigg( \dHdotdxpone \xdptwo + \dHdxpone \xddptwo + \dCdotdxpone \bigg)
    \end{split}
    \end{align} 

    \begin{align}
    \begin{split}
        & \dgammadxptwo = \HdotT \dCdxptwo + \HT \dCdotdxptwo
    \end{split}
    \end{align} 

    \begin{align}
    \begin{split}
        & \dalphaonedxpone =  \dxpone \bigg(-\dFDTdxpone \bigg) \HT \bigg(2C + 2H \xdptwo \bigg) 
        -\dFDTdxpone \bigg( 
        \dHTdxpone \bigg(2C + 2H \xdptwo \bigg)
        +
        H \bigg( 2 \dCdxpone + 2 \dHdxpone \xdptwo \bigg)
        \bigg)
    \end{split}
    \end{align} 

    \begin{align}
    \begin{split}
        & \dalphatwodxpone =  \dxpone \bigg(-\dFDTdxptwo \bigg) \HT \bigg(2C + 2H \xdptwo \bigg) 
        -\dFDTdxptwo \bigg( 
        \dHTdxpone \bigg(2C + 2H \xdptwo \bigg)
        +
        H \bigg( 2 \dCdxpone + 2 \dHdxpone \xdptwo \bigg)
        \bigg)
    \end{split}
    \end{align}

    \begin{align}
    \begin{split}
        & \dalphaonedxptwo =  \dxptwo \bigg(-\dFDTdxpone \bigg) \HT \bigg(2C + 2H \xdptwo \bigg) 
        -\dFDTdxpone 2H \dCdxptwo 
    \end{split}
    \end{align} 

    \begin{align}
    \begin{split}
    \label{dalphatwodxptwo_appendix}
        & \dalphatwodxptwo =  \dxptwo \bigg(-\dFDTdxptwo \bigg) \HT \bigg(2C + 2H \xdptwo \bigg) 
        -\dFDTdxptwo 2H \dCdxptwo 
    \end{split}
    \end{align} 

    \begin{align}
    \begin{split}
    \label{dGammadxpone_appendix}
        & \dGammadxpone = \bigg(-\dFDTdxpone \dHdxpone + (\xdptwo - \FDuzero)^T \ddHddxpone \bigg) H (\xdptwo - \FDuzero) \\
        & \hspace{1.1cm} + (\xdptwo - \FDuzero)^T \dHdxpone \bigg( \dHdxpone (\xdptwo - \FDuzero) - H \dFDdxpone \bigg)
    \end{split}
    \end{align} 

    \begin{align}
    \begin{split}
    \label{dGammadxptwo_appendix}
        & \dGammadxptwo = -\dFDTdxptwo \dHdxpone H (\xdptwo - \FDuzero) + (\xdptwo - \FDuzero)^T \dHdxpone \bigg( - H \dFDdxptwo \bigg)
    \end{split}
    \end{align} 

    The derivative of $\Omega$ wrt $[D\psX]_i$ similarly is given by
    \begin{align}
    \begin{split}
    \label{eqn: d_omega_d_Dpsi_appendix}
        \frac{d \Omega}{d [D\psX]_i}
        = \frac{d \Omega_1}{d\Xd} - (\dOmegatwodxd - \dOmegathreedxd +\dOmegafourdxd)
    \end{split}
    \end{align}

    \begin{align}
    \begin{split}
        \frac{d \Omega_1}{d \Xd}
        = 2 \begin{bmatrix} 0_{N \times N} & -I_{N \times N} \\ 
         \HTHinv \dgammadxdpone & \HTHinv \dgammadxdptwo
        \end{bmatrix}
    \end{split}
    \end{align}

    \begin{align}
    \label{eqn: d_omega2_d_psi}
    \begin{split}
         \frac{d \Omega_2}{d \Xd} = 
         \begin{bmatrix} 0_{N \times N} & \frac{1}{k}I_{N \times N}\dalphaonedxdptwo  \\ 
           0_{N \times N} &  \HTHinv \dalphatwodxdptwo) 
        \end{bmatrix} \\
    \end{split}
    \end{align}

    \begin{align}
    \label{eqn: d_omega3_d_psi}
    \begin{split}
         \frac{d \Omega_3}{d \Xd} = 
         \begin{bmatrix} 0_{N \times N} & 0_{N \times N}  \\ 
           2k \HTHinv I_{N \times N} & 0_{N \times N}
        \end{bmatrix} \\
    \end{split}
    \end{align}

    \begin{align}
    \begin{split}
        \frac{d \Omega_4}{d \Xd}
        = \begin{bmatrix} 0_{N \times N} & \frac{2}{k}I_{N \times N} \dGammadxdptwo \\ 
         0_{N \times N} & 0_{N \times N}  
        \end{bmatrix}
    \end{split}
    \end{align}

    where 
    
    \begin{align}
    \begin{split}
    \label{dalphaonedxdpone_appendix}
        \dalphaonedxdptwo = -\dFDTdxpone 2 \HT H
    \end{split}
    \end{align}

    \begin{align}
    \begin{split}
    \label{dalphaonedxdptwo_appendix}
        \dalphatwodxdptwo = -\dFDTdxptwo  2 \HT H
    \end{split}
    \end{align}
    
    \begin{align}
    \begin{split}
    \label{dgammaxdpone_appendix}
        \dgammadxdpone
        =  \dHdotTdxdpone \bigg(H \xdptwo + C \bigg) + H^T \bigg( \dHdotdxdpone \xdptwo + \dCdotdxdpone \bigg)
    \end{split}
    \end{align}

    \begin{align}
    \begin{split}
    \label{dgammaxdptwo_appendix}
        \dgammadxdptwo = (\HdotT H + \HT \Hdot) + \HT \dCdotdxdptwo
    \end{split}
    \end{align}

    \begin{align}
    \begin{split}
    \label{dGammadxdptwo_appendix}
        & \dGammadxdptwo = I_{N \times N} \dHdxpone H (\xdptwo - \FDuzero) + (\xdptwo - \FDuzero)^T \dHdxpone H
    \end{split}
    \end{align} 

   $\dgammadxdpone$ \eqref{dgammaxdpone_appendix} and $\dgammadxdptwo$ \eqref{dgammaxdptwo_appendix} can be simplified even further.
   Next, we highlight the following relations that will be used to simplify these terms.
   First, $\Hdot$ can be computed using the chain rule in the following way:

    \begin{align}
    \begin{split}
        \Hdot
        =  \dHdxpone \frac{\partial \xpone}{\partial t} = \dHdxpone \xdpone
    \end{split}
    \end{align}

    Second, $\Cdot$ can also be computed similarly using the chain rule:

    \begin{align}
    \begin{split}
        \Cdot = \dCdxpone \xdpone + \dCdxptwo \xdptwo
    \end{split}
    \end{align}
    
    Lastly, recall that $H$ is symmetric, therefore:
    \begin{align}
    \begin{split}
        \dHdxpone =  \dHTdxpone
    \end{split}
    \end{align}
    
    Using these three relations we can obtain the following: 
    \begin{align}
    \label{dHdotdxdpone_appendix}
    \begin{split}
        \dHdotdxdpone =  \dHdxpone
    \end{split}
    \end{align}

    \begin{align}
    \label{dCdotdxdpone_appendix}
    \begin{split}
        \dCdotdxdpone =  \dCdxpone
    \end{split}
    \end{align}

    Using \eqref{dHdotdxdpone_appendix} and \eqref{dCdotdxdpone_appendix} to simplify $\dgammadxdpone$ \eqref{dgammaxdpone_appendix} yields:
    \begin{align}
    \begin{split}
    \label{dgammaxdpone_simp_appendix}
        \dgammadxdpone
        =  \dHdxpone \bigg(H \xdptwo + C \bigg) + H \bigg( \dHdxpone \xdptwo + \dCdxpone \bigg)
    \end{split}
    \end{align}

    we can apply similar line of reasoning to obtain a simplification for $\dgammadxdptwo$ \eqref{dgammaxdptwo_appendix}
    \begin{align}
    \begin{split}
    \label{dgammaxdptwo_simp_appendix}
        \dgammadxdptwo
        =  \Hdot H + H (\Hdot + \dCdxptwo)
    \end{split}
    \end{align}

    Lastly, the derivative of $\Omega$ wrt $[D^2\psX]_i$ is given by

    \begin{align}
    \begin{split}
        \frac{d \Omega}{d [D^2\psX]_i}
        = 2\begin{bmatrix} I_{N \times N} & 0_{N \times N} \\ 
         0_{N \times N} & \HTHinv \dgammadxddptwo 
        \end{bmatrix}
    \end{split}
    \end{align}

    where 

    \begin{align}
    \begin{split}
        \dgammadxddptwo = \HT H
    \end{split}
    \end{align}

    This simplifies to 
    \begin{align}
    \begin{split}
    \label{eqn: d_omega_dd_psi_appendix}
        \frac{d \Omega}{d [D^2\psX]_i}
        = 2\begin{bmatrix} I_{N \times N} & 0_{N \times N} \\ 
         0_{N \times N} & I_{N \times N} 
        \end{bmatrix}
    \end{split}
    \end{align}

    Observing the terms from Equations \eqref{gamma_appendix} -- \eqref{dGammadxptwo_appendix}, we have that $\frac{d \Omega}{d \psX_i}$ \eqref{eqn: d_omega_d_psi_appendix} is a function of the rigid body dynamics and its higher-order derivatives.
    Namely,
    
    \begin{align*}
    \begin{split}
         \frac{d \Omega}{d \psX_i} \bigg(H, \Hdot, C, \Cdot, \FDuzero, \dHdxpone, \ddHddxpone, \dHdotdxpone, \dmHinvCdxpone, \dmHinvCdxptwo, \dCdxpone, \dCdxptwo, \dCdotdxpone, \dCdotdxptwo, \ddFDddxpone, \ddFDddxptwo, \ddFDdxponetwo \bigg)
    \end{split}
    \end{align*} 

    Similarly, observing the terms from Equations \eqref{dalphaonedxdpone_appendix},
    \eqref{dalphaonedxdptwo_appendix}, \eqref{dGammadxdptwo_appendix},
    \eqref{dgammaxdpone_simp_appendix}
    and 
    \eqref{dgammaxdptwo_simp_appendix}
    , we have that $\frac{d \Omega}{d [D\psX]_i}$ \eqref{eqn: d_omega_d_Dpsi_appendix} is a function of the rigid body dynamics and its first-order derivatives.
    Specifically,
    
    \begin{align*}
    \begin{split}
        \frac{d \Omega}{d [D\psX]_i} \bigg(H, C, \Hdot, \FDuzero, \dHdxpone, \dmHinvCdxpone, \dmHinvCdxptwo, \dCdxpone, \dCdxptwo \bigg)
    \end{split}
    \end{align*}

    and from \eqref{eqn: d_omega_dd_psi_appendix}, we have 
    
    \begin{align*}
    \begin{split}
        \frac{d \Omega}{d [D^2\psX]_i}
        = 2\begin{bmatrix} I_{N \times N} & 0_{N \times N} \\ 
         0_{N \times N} & I_{N \times N} 
        \end{bmatrix}
    \end{split}
    \end{align*}

\end{proof}
\section{Results Appendix}
\label{subsec: results_appendix}
This section contains all the varied parameters that we grid search over and use for the experiments in Section \ref{sec:experiments}: 

\begin{table}[h]
    \centering
    \begin{tabular}{|l|c|}
    \hline
    \textbf{Parameter}    & \textbf{Grid values} \\ \hline
    $w_u$                 & $[10^{-4}, 10^{-3}, 10^{-2}, 10^{-1}]$ \\ \hline
    $w_x$                 & $[10^{-4}, 10^{-3}, 10^{-2}, 10^{-1}, 0.0, 1, 10]$ \\ \hline
    $w_{xf}$              & $[10^{-4}, 1.0, 10, 1000]$ \\ \hline
    $\delta_t$                  & $[10^{-2}, 10^{-3}, 10^{-4}]$ \\ \hline
    \textit{Initial guess} & [Zeros, Line and RNEA, Constant] \\ \hline
    \end{tabular}
    \caption{Parameter Values for Crocoddyl unconstrained}
    \label{table: Crocoddyl grid without obstacles}
\end{table}

\begin{table}[h]
    \centering
    \begin{tabular}{|l|c|}
    \hline
    \textbf{Parameter}    & \textbf{Grid values} \\ \hline
    $w_u$                 & $[10^{-5}, 10^{-4}, 10^{-3}, 10^{-2}, 10^{-1}]$ \\ \hline
    $w_x$                 & $[10^{-6}, 10^{-4}, 10^{-3}, 0.0, 1, 100, 1000]$ \\ \hline
    $w_{xf}$              & $[10^{-6}, 10^{-4}, 1.0, 10, 1000]$ \\ \hline
    $\delta_t$                  & $[10^{-2}, 10^{-3}]$ \\ \hline
    $\epsilon_{tol}$                 & $[10^{-3}, 10^{-4}, 10^{-7}]$ \\ \hline
    $\mu_{init}$          & $[10^{-2}, 10^{-7}, 10^{-8}]$ \\ \hline
    \textit{Initial guess} & [Zeros, Line and RNEA, Rollout and Zero, Rollout and Constant]\\ \hline
    \end{tabular}
    \caption{Parameter Values for Aligator with obstacles}
    \label{table: Aligator grid with obstacles}
\end{table}

\begin{table}[h]
    \centering
    \begin{tabular}{|l|c|}
    \hline
    \textbf{Parameter}    & \textbf{Grid values} \\ \hline
    $s_{max}$             & $[10^{-4}, 10^{-3}, 10^{-2}, 10^{-1}, 1, 5, 10, 25, 50, 100]$ \\ \hline
    $k$                   & $[10^{3}, 10^{4}, 10^{5}, 10^{6}, 10^{7}, 10^{8}, 10^{9}, 10^{10}]$ \\ \hline
    $p$                   & $[5, 6, 7, 8, 9, 10, 15]$ \\ \hline
    \textit{Initial guess} & [Line] \\ \hline
    \end{tabular}
    \caption{Parameter Values for  \methodname{} for pendulum swingup}
    \label{table: aghf grid pendulum swingup}
\end{table}

\begin{table}[h]
    \centering
    \begin{tabular}{|l|c|}
    \hline
    \textbf{Parameter}    & \textbf{Grid values} \\ \hline
    $s_{max}$             & $[10^{-4}, 10^{-3}, 10^{-2}, 10^{-1}, 1, 2, 3, 4, 5, 6, 7, 8, 9, 10, 25, 50, 100]$ \\ \hline
    $k$                   & $[10^{4}, 10^{5}, 10^{6}, 10^{7}, 10^{8}, 10^{9}, 10^{10}]$ \\ \hline
    $p$                   & $[5, 6, 7, 8, 9, 10, 15]$ \\ \hline
    \textit{Initial guess} & [Line] \\ \hline
    \end{tabular}
    \caption{Parameter Values for \methodname{} for Kinova without obstacles}
    \label{table: aghf grid Kinova without obstacles}
\end{table}

\begin{table}[H]
    \centering
    \begin{tabular}{|l|c|}
    \hline
    \textbf{Parameter}    & \textbf{Grid values} \\ \hline
    $s_{max}$             & $[10^{-4}, 10^{-3}, 10^{-2}, 10^{-1}, 1, 2, 3, 4, 5, 6, 7, 8, 9, 10, 25, 50, 100]$ \\ \hline
    $k$                   & $[10^{3}, 10^{4}, 10^{5}, 10^{6}, 10^{7}, 10^{8}, 10^{9}, 10^{10}]$ \\ \hline
    $p$                   & $[5, 6, 7, 8, 9, 10]$ \\ \hline
    \textit{Initial guess} & [Line] \\ \hline
    \end{tabular}
    \caption{Parameter Values for \methodname{} for Digit without obstacles}
    \label{table: aghf grid Digit without obstacles}
\end{table}

\begin{table}[H]
    \centering
    \begin{tabular}{|l|c|}
    \hline
    \textbf{Parameter}    & \textbf{Grid values} \\ \hline
    $s_{max}$             & $[10^{-4}, 10^{-2}, 10^{-1}, 1, 5, 10, 25, 50, 100]$ \\ \hline
    $k$                   & $[10^{5}, 10^{6}, 10^{7}, 10^{8}, 10^{9}, 10^{10}]$ \\ \hline
    $p$                   & $[5, 6, 7, 8, 9, 10, 15]$ \\ \hline
    $\ccons$      & $[1, 50, 200]$ \\ \hline
    $k_{\text{cons}}$      & $[10^{5}, 10^{6}, 10^{7}, 10^{8}, 10^{9}, 10^{10}]$ \\ \hline
    \textit{Initial guess} & [Line] \\ \hline
    \end{tabular}
    \caption{Parameter Values for \methodname{} Kinova with Constraints}
    \label{table: aghf grid kinova with constraints}
\end{table}

\begin{table}[h]
    \centering
    \begin{tabular}{|l|c|}
    \hline
    \textbf{Parameter}    & \textbf{Grid values} \\ \hline
    $s_{max}$             & $[10^{-4}, 10^{-2}, 10^{-1}, 1, 5, 10, 25, 50, 100]$ \\ \hline
    $k$                   & $[10^{5}, 10^{6}, 10^{7}, 10^{8}, 10^{9}, 10^{10}]$ \\ \hline
    $p$                   & $[5, 6, 7, 8, 9, 10, 15]$ \\ \hline
    $\ccons$      & $[1, 50, 200]$ \\ \hline
    $k_{\text{cons}}$      & $[10^{5}, 10^{6}, 10^{7}, 10^{8}, 10^{9}, 10^{10}]$ \\ \hline
    \textit{Initial guess} & [Line] \\ \hline
    \end{tabular}
    \caption{Parameter Values for \methodname{} for Digit with Constraints}
    \label{table: aghf digit grid with constraints}
\end{table}

\end{document}